\documentclass[review, times, authoryear]{elsarticle}
\usepackage[bindingoffset=0.2in,%
           left=1in,right=1in,top=1in,bottom=1in]{geometry}
\usepackage{amsmath}
\usepackage{dsfont}
\usepackage{listings}
\usepackage{color}
\usepackage{bold-extra}
\usepackage{booktabs}
\usepackage{graphicx}
\usepackage{adjustbox}
\usepackage{afterpage}
\usepackage{epstopdf}
\usepackage{rotating}
\usepackage{graphics}
\usepackage{xcolor}
\usepackage[ruled,linesnumbered]{algorithm2e}
\usepackage{xspace}
\usepackage{pdflscape}

\renewcommand{\arraystretch}{0.8}

\usepackage{chngcntr}
\definecolor{light-gray}{gray}{0.95}
\usepackage{tikz}
\allowdisplaybreaks
\usetikzlibrary{calc, math, positioning, arrows}

\usepackage[hidelinks]{hyperref}

\usepackage{hyperref, color}

\begin{document}
\bibliographystyle{elsarticle-harv}
\begin{frontmatter}

\title{Integration of returns and decomposition of customer orders in e-commerce warehouses}

\author[a]{Albert H. Schrotenboer\corref{mycorrespondingauthor}}
\ead{a.h.schrotenboer@rug.nl} 
\cortext[mycorrespondingauthor]{Corresponding author. Nettelbosje 2, 9747 AE Groningen, The Netherlands.}
\author[b]{Susanne Wruck}
\author[a]{Iris F. A. Vis}
\author[a]{Kees Jan Roodbergen}

\address[a]{Department of Operations, Faculty of Economics and Business, University of Groningen, Groningen, The Netherlands}
\address[b]{Flaconi GmbH, Berlin, Germany}

\begin{abstract}
In picker-to-parts warehouses, order picking is a cost- and labor-intensive operation that must be designed efficiently. It comprises the construction of order batches and the associated order picker routes, and the assignment and sequencing of those batches to multiple order pickers. The ever-increasing competitiveness among e-commerce companies has made the joint optimization of this order picking process inevitable. Inspired by the large number of product returns and the many but small-sized customer orders, we address a new integrated order picking process problem. We integrate the restocking of returned products into regular order picking routes and we allow for the decomposition of customer orders so that multiple batches may contain products from the same customer order. We thereby generalize the existing models on order picking processing. We provide Mixed Integer Programming (MIP) formulations and a tailored adaptive large neighborhood search heuristic that, amongst others, exploits these MIPs. We propose a new set of practically-sized benchmark instances, consisting of up to 5547 to be picked products and 2491 to be restocked products. On those large-scale instances, we show that integrating the restocking of returned products into regular order picker routes results in cost-savings of 10 to 15\%. Allowing for the decomposition of the customer orders' products results in cost savings of up to 44\% compared to not allowing this. Finally, we show that on average cost-savings of 17.4\% can be obtained by using our ALNS instead of heuristics typically used in practice.
\end{abstract}


\end{frontmatter}

\section{Introduction}

Technological innovations, high competition, and increasing service requirements necessitate the design and maintenance of efficient customer order processing, preferably with short response times and minimal effort costs \citep{xu2009, boysen2018warehousing}. For the warehouses of leading e-commerce companies such as Alibaba, Amazon or JD, efficient customer order processing reveals the need to adapt order picking activities to be able to handle a large number of typically small-sized customer orders \citep[][]{boysen2018manual}. Besides, e-commerce warehouses face a considerable number of product returns, which must be restocked into the warehouse and create additional labor demands \citep{koster2002, mostard2005, su2009, schrotenboer2017order}. 
 In this paper, we study a comprehensive order processing problem to provide insights on how picker-to-parts warehouse operations are impacted by these developments. That is, we consider a joint order-picker batching, assignment, sequencing and routing model \citep[see, e.g.,][]{scholz2017order}, in which we exploit 1) the small-sized customer orders by allowing the decomposition of customer orders so that the products of a customer order may be picked in multiple batches, and 2) we integrate the restocking of return products in the regular order picking operations. We refer to this practically relevant problem as the Generalized Joint Order Batching, Assignment, Sequencing and Routing Problem (G-JOBASRP).



The two distinct features that we incorporate in current warehouse operations aim to increase its flexibility. First, it is well-known that splitting up customer orders within distribution networks reduces the transportation costs in the complete supply chain \citep{zhang2019multi}. In the context of warehouse operations, separating customer orders (i.e., the contained order lines) amongst multiple order picking batches may similarly increase the flexibility of the order picking process. Namely, travel times (within the warehouse) can be shortened leading to a larger warehouse capacity to process customer orders. This may especially be true for e-commerce businesses that face large numbers of relatively small customer orders. We, therefore, investigate if allowing for splitting up customer orders among multiple batches has a cost-saving potential large enough to compensate for the inevitable additional handling operations.

Such additional handling operations due to splitting up customer orders amongst multiple batches are either sorting operations further downstream within the warehouse or additional operations in the last-mile due to multiple shipments to the same customer. However, the split-up of customer orders within picker-to-parts warehouses has not been investigated in the literature. Hence, the trade-off between the gained efficiency and the required additional handling operations is not known. The model that we present aims to make this quantification by the inclusion of an order split-up cost that represents the costs of additional handling operations in a unified way. That is, the split-up costs can be interpreted as the costs of recollecting the split-up order within the warehouse or as the extra shipping costs incurred by sending multiple shipments to the same customer.

The second distinct feature we add is the integration of the restocking of returned products in the regular customer order processing operations. Prior research has shown that the efficiency of the order picking processes is increased by integrating individual processes. For instance, integrating order-picker routing and batching  \citep{valle2017optimally}, as well as integrating the assignment and sequencing of those batches to multiple order pickers \citep{henn2015order, scholz2017order}. Jointly studying such  processes is also advocated by the recent review on warehouse systems by \cite{boysen2018warehousing}. We, therefore, investigate whether the recognized cost-saving potential of incorporating the restocking of returned products in case of a single order picker, as detailed in \cite{schrotenboer2017order}, also suffices in comprehensive  environments that include order batching, sequencing and assignment. 

The integration of both distinct features might increase the efficiency of warehouse operations even further. However, the operational circumstances under which this integration is most profitable are not  known. For instance, the split-up of customer orders will simplify the actual routing of order pickers, since batches can consist of products that are in relative proximity of each other. This may lead to a reduction in travel costs (inside the warehouse) and thereby simplifies the integration of the restocking of returned products. On the other hand, the complexity of the order picker routing and its associated travel costs will increase due to the presence of customer order deadlines and the stylized split-up costs. The construction of efficient order picker routes will  be a balancing act between the travel costs inside the warehouse on the one hand, and the extent to wich orders are split-up and the associated split-up costs incurred outside our scope. This increased complexity might reduce the efficiency of integrating product returns in the regular order picking operations. Hence, experiments and insights are needed in order to accomplish a successful implementation of such integrated policies in future warehouses.

In this paper, we generalize the Joint Order Batching, Assignment, Sequencing, and Routing problem (JOBASRP) introduced by \cite{scholz2017order}. We incorporate the restocking of returned products into regular order picking routes and allowing customer orders to be split-up at the price of incurring split-up costs. 
For sufficiently high split-up costs, all order lines of a customer order are assigned to a single batch. Setting the split-up costs to zero equals handling each order line of a customer order individually. We refer to our problem as the Generalized Joint Order Batching, Assignment, Sequencing, and Routing problem (G-JOBASRP). Its goal is to find a cost-minimizing solution for processing a pool of customer orders and product returns with multiple order pickers. Each customer order has a specific deadline before it needs to be processed. The solution determines, for each order picker, a sequence of batches consisting of the products to be picked and returned, while taking into account the actual order picker routes of each batch.

We formalize the G-JOBASRP by developing a compact Mixed Integer Programming (MIP) formulation, which is independent of the actual warehouse layout. We also provide two additional MIP formulations that result from the compact MIP formulation by applying two different Dantzig-Wolfe reformulations. We aim to solve practically sized instances, which typically consists of 1000's of order lines per day for a warehouse \citep{koster2002}. Therefore, we develop a tailored  Adaptive Large Neighborhood Search (ALNS) heuristic \citep[see, e.g., ][]{ropke2006adaptive}. It consists of a carefully selected set of operators that build upon concepts from ALNS, Iterative Local Search  \citep[see, e.g.,][]{lourencco2003iterated}, and uses the extended MIP formulations. 

Unfortunately, the benchmark instances of \cite{scholz2017order} are, after contacting the authors, not available anymore. We, therefore, provide a new set of practically inspired benchmark instances that are publically available \footnote{www.albertschrotenboer.com}. The benchmark instances consist of up to 5547 to be picked order lines and 2491 to be restocked products. On those instances, the ALNS provides on average 17.4\% percent better than an intuitive and practically often observed heuristic. Besides, those instances show that incorporating the restocking of returned products into regular order picker routes results in cost-savings of up to 19\%.

Moreover, we analyze  the effect of splitting up customer orders among multiple batches at the price of incurring additional split up costs. Maximum cost savings around 45\% are obtained due to the increased flexibility while creating order picking batches. Besides, we show that the impact of split-up costs is indirectly observed by an increase in the travel costs associated with picking orders inside the warehouse, as customer order split-ups will then typically be avoided.

The remainder of the paper is structured as follows. We provide a review of selected literature in Section \ref{sec:literature} and present the mixed integer programming formulations of the G-JOBASRP in Section \ref{sec:problem}. In Section \ref{sec:approach}, we describe the adaptive large neighborhood search heuristic in detail. In Section \ref{sec:numericalresults}, we summarize and discuss our numerical results and we conclude and provide suggestions for future research in Section \ref{sec:conclusions}.

\section{Literature Review}
\label{sec:literature}
Warehouse operations, including order picking processes, have received substantial research attention \citep{rouwenhorst2000, gu2007, koster2007, roodbergen2009, gong2011, marchet2015investigating, staudt2015warehouse, davarzani2015toward, van2017designing}.  \cite{van2017designing} offer an extensive overview of research on combining planning problems within warehouse operations. On an operational level, they urge researchers to continue the studying of integrated decision making, and advise the further incorporation of aspects related to e-commerce and globalization. We believe that the G-JOBASRP follows this spirit of incorporating e-commerce aspects in integrated decision making. Namely, we incorporate the restocking of returned products into traditional order picking routes, and we allow customer orders to be split up amongst order picking batches. In the following, we review selected work being relevant or related to the G-JOBASRP. We first review work on (integrated) batching and routing, which is a subproblem of major relevance for the G-JOBASRP. Although its stems from the early 20000's, we judge the discussion to be essential since it forms the basis for recent papers on integrated decision problems within the order picking process. 

When order picker routing is the focal problem, the underlying assumption is that the batch (i.e., the set of locations to be visited) is known beforehand. In this case, the problem forms a special case of the traveling salesman problem. \cite{ratliff1983} were the first to design an optimal solution approach for the case of a rectangular warehouse with parallel aisles and two cross aisles. \cite{roodbergen2001b} extend their approach for warehouses with a middle cross-aisle. However, product returns make an extension of these optimal approaches to more general warehouse situations cumbersome. Hence, also the warehouse routing problem has been approached mostly with heuristic methods. For example, \cite{roodbergen2001} provide a discussion of several routing heuristics for warehouses with more cross aisles. The authors also propose the combined and combined+ heuristics and demonstrate that the latter yields good results compared with S-shape and largest gap routing. \cite{theys2010} also consider routing methods in multi-parallel-aisle warehouses and opportunities to decrease travel distances by using meta-heuristic approaches rather than constructive heuristics. Recently, a metaheuristic approach is developed in \cite{schrotenboer2017order} for solving the order picker routing problem with product returns. Their work shows that incorporating the restocking of product returns in the regular order picking routes results in substantial cost-savings. To the best of the authors' knowledge, it is the only work that incorporated the restocking of return products in regular order picking routes. In this paper, we investigate its cost-savings potential in the richer setting of the G-JOBASRP.

Whereas the order picker routing literature assumes predetermined batches, the order batching literature typically assumes fixed routing policies (e.g., S-shape or largest-gap). The design of such order batching policies usually relies on heuristic approaches because of the high combinatorial complexity of this problem. \cite{gademann2001} prove that the batching problem is NP-hard in the strong sense when the number of products in a  batch is greater than two. \cite{gademann2005} use a branch-and-price algorithm to solve the batching problem for up to $300$ order lines. The batch size, however, was limited to $10$ order lines in their experiments, which might be restrictive for large warehouses. \cite{chen2005} and \cite{ho2008} show that that there is a large potential for sophisticated (metaheuristic) search methods.  The so-called savings heuristics form another stream of solution methods for solving the batching problem. They rely on iteratively improving the solution, for instance, by merging batches until no improvement can be found. \cite{koster1999} provide an overview of such methods and also show that they substantially outperform myopic policies. The integration of product returns in order batching is first discussed by \cite{wruck2012}. In their work, the authors considered this integrated process while assuming fixed routing policies. To the best of the author's knowledge, this is the only work in the batching literature that considers product returns and customer orders simultaneously. It is, therefore, interesting to study this in the richer environment of the G-JOBASRP. 

The ever-increasing competitiveness of e-commerce companies causes an increased interest in integrated decision making within the order picking process.  \cite{tsai2008} offer a first step toward the integration of the batching and routing problem for a specific warehouse layout by developing a genetic algorithm (GA). To find an efficient batch partitioning, they use an outer GA to identify the batch formation, and then an inner GA to evaluate the quality of these batches by optimizing the route. An exact algorithm for solving the joint order picker routing and batching problem has recently been proposed by \cite{valle2017optimally}.

In \cite{henn2013metaheuristics}, the joint order batching and sequencing problem is studied that penalizes both travel costs and tardiness (collecting customer orders after their corresponding deadline). Metaheuristic solution approaches have shown that cost-savings up to 46\% can be attained compared to constructive heuristics. An improvement upon this work is presented by \cite{menendez2017general}. Multiple order pickers are then considered by \cite{henn2015order}, and the authors solve the resulting optimization problem with a variable neighborhood search. Instead of considering multiple order pickers,  \cite{chen2015efficient} consider a generalization of the joint order batching and sequencing problem by including the optimization of order picker routes. \cite{scholz2017order} then provide a unified generalization, with both multiple pickers and order picker routing, what they call the Joint Order Batching, Sequencing, Assignment, Routing Problem (JOBASRP). The authors propose a metaheuristic solution procedure and use it to obtain insights into the resulting efficiency of joint optimization. In this paper, we intend to improve this efficiency even further by accounting for two generalizing aspects. First, we include the restocking of return products, which is already shown to be efficient for subproblems of the JOBASRP \citep{wruck2012, schrotenboer2017order}. Second, we allow customer orders to be split up amongst batches at the price of a split-up cost. We are not aware of any studies that studied the latter trade-off between splitting up customer orders and the resulting cost increase. 


\section{Problem Definition}

\begin{figure*}
\begin{minipage}{0.45\linewidth}
\centering 
\small
\textbf{To be picked:}
\begin{tabular}{lrrr}
\toprule
Cust. & Deadline & order line(s) & assigned batch \\  \midrule
1     & 10:15 & 6 $\times$ i-23 & \textbf{A} \\
      & 10:15 & 2 $\times$ i-24 & \textbf{A} \\
      & 10:15 & 1 $\times$ i-08 & \textbf{B} \\
      & 10:15 & 9 $\times$ i-26 & \textbf{A} \\
2     & 9:45 & 1 $\times$ i-09 & \textbf{B} \\
      & 9:45 & 3 $\times$ i-04 & \textbf{B} \\
\ldots & \ldots & \ldots & \ldots \\
8   & 11:00 & 6 $\times$ i-98 & \textbf{I} \\
8   & 11:00 & 3 $\times$ i-22 & \textbf{F} \\
8   & 11:00 & 4 $\times$ i-46 & \textbf{F} \\
\end{tabular}
\end{minipage}
\begin{minipage}{0.5\linewidth}
\centering 
\small
\textbf{To be returned:}\\
\begin{tabular}{lr}
\toprule
order line(s) & assigned batch \\  \midrule
1 $\times$ i-09 & \textbf{E} \\
2 $\times$ i-56 & \textbf{C} \\
1 $\times$ i-49 & \textbf{A} \\ \bottomrule
\end{tabular}
\begin{tikzpicture}[scale = 0.9, baseline=(current bounding box.center)]
\draw  (-3.5,3.5) rectangle (-1,3);

\node at (-4.5,3.25) {1};
\node at (-4.5,3.75) {Picker};
\node (v1) at (-5,3.5) {};
\node (v2) at (-4,3.5) {};
\draw  (v1) edge (v2);
\draw  (-0.75,3.5) rectangle (0.5,3);
\draw  (-2,2.75) rectangle (0.25,2.25);
\draw  (-3.5,2.75) rectangle (-2.25,2.25);
\draw  (0.5,2.75) rectangle (1.25,2.25);
\draw  (0.75,3.5) rectangle (3.25,3);
\draw  (1.5,2.75) rectangle (3,2.25);
\draw  (-3.5,2) rectangle (-1.5,1.5);

\draw  (-1.25,2) rectangle (-0.25,1.5);
\draw  (1.5,1.75) rectangle (1.5,1.75);
\draw  (0,2) rectangle (3.75,1.5);
\node at (-4.5,2.5) {2};
\node at (-4.5,1.75) {3};
\node (v5) at (-3.5,4.5) {09:00};
\node (v7) at (-0.25,4.5) {10:00};
\node (v8) at (-0.25,3.75) {};

\node (v3) at (-3.5,3.75) {};
\node (v4) at (3.25,3.75) {};
\node (v6) at (3.25,4.5) {11:00};
\draw  (v3) edge (v4);
\draw [-triangle 60] (v3) edge (v5);
\draw [-triangle 60] (v4) edge (v6);
\draw [-triangle 60] (v8) edge (v7);

\node at (-2.25,3.25) {\textbf{A}};
\node at (-0.15,3.25) {\textbf{E}};
\node at (2, 3.25) {\textbf{F}};

\node at (-2.85,2.50) {\textbf{B}};
\node at (-1,2.50) {\textbf{C}};
\node at (0.9,2.50) {\textbf{G}};
\node at (2.25, 2.50) {\textbf{I}};

\node at (-2.25, 1.75) {\textbf{H}};
\node at (-0.75, 1.75) {\textbf{J}};
\node at (2, 1.75) {\textbf{D}};
\end{tikzpicture}
\end{minipage}

\caption{Illustrative example of the G-JOBSARP with three order pickers. On the left, a list of customer orders consisting of multiple order lines and the assigned batches. On the right, a possible schedule of batches to be processed by the order pickers. The processing time of each batch depends on the order picker route within the batch.}
\label{fig:example}
\end{figure*}
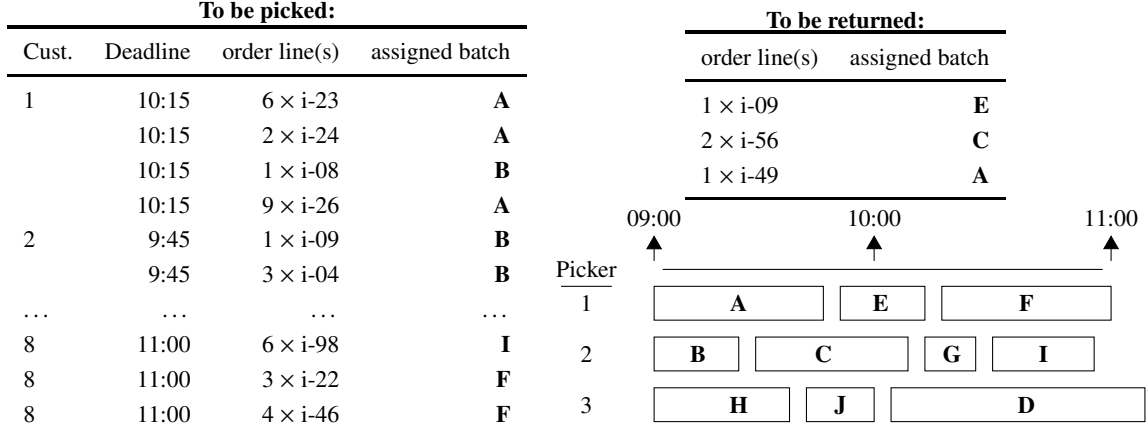

\label{sec:problem}
In this section, we present a Mixed Integer Programming (MIP) formulation of our Generalized Joint Order Batching, Assignment, Sequencing and Routing Problem (G-JOBASRP). 

In the following, we consider \textit{customer orders} that consists out of multiple \textit{order lines}. Each order line consists of a number of the same \textit{products}. A product is assumed to be stored at a single location in the warehouse. The weight of an order line equals the weight of the associated product weight multiplied with the number of times the product is ordered on that order line. We assume that each order line is picked by at most a single order picker, i.e., the products associated to a single order line (that are all equal) are picked in the same batch. We allow the order lines associated to the same customer order to be split-up amongst multiple batches, but incur split-up costs if we do so. To keep notation concise, products that need to be restocked in the warehouse are defined as customer orders of a single order line (possibly consisting of multiple products) with a negative weight.

Customer order deadlines apply to all order lines contained. The deadlines for order lines consisting of returned products are set to occur at the end of the time horizon. The transport capacity of an order picker is limited and cannot be exceeded at any point of a route. Note that due to the inclusion of product returns it is required to keep track of the order picker's capacity along the route. We assume that the travel speed of all order pickers and the time required to pick or return an order line are constant. 

The goal of the G-JOBASRP is then to process all the order lines by creating order picking batches and its associated routing, as well as to assign those create batches to an order picker and sequence the assigned batches for each order picker. The objective consists of travel costs (the routing of the batches), tardiness costs incurred if an customer order deadline is exceeded, and split-up costs (representing costs incurred outside our operations) due to splitting the order lines of the same customer amongst multiple batches.

The MIP formulation of the G-JOBASRP is independent of most specific warehouse characteristics, including the layout; only the distances between locations need to be specified. We consider a single depot where all routes of the pickers begin and end. All  order lines picked are delivered to the depot and all order lines returned are picked from the depot. The number of available order pickers is given. We detail our notations for the main parameters and decision variables in Tables \ref{parameters} and \ref{variables}.

Before detailing the problem mathematically, let us illustrate the main characteristics and terminology used. We provide an illustrative example of the G-JOBASRP in Figure \ref{fig:example}. Here, three elements of the G-JOBASRP are presented. On the left, we see a list of \textit{customer orders} (Cust.), with associated \textit{order lines} to be picked (e.g., 6 times product `i-23'). We assign these order lines to different batches (called A, B, $\ldots,$ F). It is not allowed to assign, for example, the six times product `i-23' (order line 1) to multiple batches. On the top right, we see several order lines that require restocking (e.g., one times product `i-09'). A solution to this example is provided on the bottom right, where we scheduled the batches A until F among three available order pickers. The length of each batch in this schedule is determined by the actual route the order picker traverses through the warehouse to process the assigned batches.



\begin{table}
\centering
\caption{Overview of the main parameters of the G-JOBASRP}

\begin{tabular}{ll|ll} 
\toprule
$\mathcal{C} = \{0, \ldots, C \}$    & Set of all customer orders 	 	& $w_i$		& Weight of one product in order line $i$\\
$\mathcal{I} = \{1, \ldots, I \}$    & Set of all order lines 	 	& $v$ 			& Travel speed \\
$\mathcal{I}^c $ & Order lines of customer order $c$ & $\beta$	& Customer order split-up costs \\ 
$\mathcal{E} = \{1, \ldots , E\}$    & Set of order pickers		 	& $s_{\text{break}}$ 	& Break length between routes \\ 
$\mathcal{H}^e = \{1, \ldots, H^e\}$ & Set of batches of picker $e$ 	& $s_{\text{pick}}$ 	& Pick time per product  \\ 
$\hat{\mathcal{B}}$ & Set of assigned routes   & $s$ 			& Travel cost per time unit  \\ 
$\hat{\mathcal{S}}$ & Set of assigned schedules & $l_i$ & Number of products of order line $i$ \\ 
$q^\textsc{p}_i$ 						 & Quantity of $i \in \cal{I}$ to pick   & $d_{n_1n_2}$	& Distance between $n_1,n_2 \in \mathcal{N}$  \\
$q^\textsc{r}_i$ 						 & Quantity of $i$ to return 	& $\alpha$		& Delay penalty per product and time unit \\
$\bar{d}_i$					   		 & Deadline of customer order $c$ & Q			& Capacity of picking device \\
$T$								 & Time horizon length		&  & \\
\bottomrule
\end{tabular}
\label{parameters}
\end{table}

\begin{table}
\centering
\caption{Overview of decision variables and accompanying parameters}
\begin{tabular}{ll} 
\toprule
\multicolumn{2}{l}{Compact MIP Formulation (Section 3.1)}\\ \midrule
$b_i^{he} \in\{0,1\}$				& Equals $1$ if order line $i \in \cal{I}$ is contained in the $h$-th batch of picker $e$ \\
$\tilde{b}_c^{he} \in\{0,1\}$				& Equals $1$ if order line $i$ is contained in the $h$-th batch of picker $e$ \\
$\chi^{he}_{ij} \in\{0,1\}$ 		& Equals $1$ if order line $j$ is ``visited" after order line $i$ in the $h$-th batch of picker $e$ \\
$\psi^{he}_{ij} \in \mathbb{R}_+$	& Total load of already picked order lines, transported along arc $(i,j)$ \\
$\omega^{he}_{ij} \in \mathbb{R}_+$ & Total load of order lines to be delivered, transported along this arc $(i,j)$ \\
$\textsc{td}^{he} \in \mathbb{R}_+$			& Travel distance of the route of the $h$-th batch of picker $e$ \\
$\tau_i \in \mathbb{R}_+$			& Tardiness of order line $i$ \\
$c_i \in \mathbb{R}_+$				& Completion time of order line $i$ \\
$c_i^{eh} \in \mathbb{R}_+$  & Modeling variable; completion time of order line $i$ of picker $e$ if it is contained in batch $h$, 0 otherwise \\
$\kappa^1_{ihe}, \ \kappa^2_{ihe} \in \mathbb{R}_+$  & Modeling variables for linearizing purposes \\
$\textsc{st}^{he}, \textsc{co}^{he} \in \mathbb{R}_+ $ 	& Start and completion time of batch $h$ of picker $e$ \\ \midrule
\multicolumn{2}{l}{Extended MIP formulations (Section 3.2)} \\ \midrule
$x^{\hat{b}}_{eh} \in \{0, 1\}$ & Variable equals 1 if $\hat{b}$ is assigned to batch $h$ of picker $e$\\ 
$\xi^{\hat{b}}_i \in \{0, 1\}$  & Parameter equals 1 if $i$ is in batch $\hat{b} \in \hat{\cal{B}}$ \\
$y^{\hat{s}} \in \{0, 1\}$ & Equal to 1 if schedule $\hat{s}\in\hat{\cal{S}}$ is selected \\
$\eta^{\hat{s}}_c \in \mathbb{N}_+$  & The number of distinct batches with products of customer $c$ \\
$\xi^{\hat{s}}_i \in \{0, 1\}$  & Parameter equals 1 if $i$ is in schedule $\hat{s} \in \hat{\cal{S}}$. \\

\bottomrule
\end{tabular}
\label{variables}
\end{table}
\subsection{Mixed Integer Programming Formulation}
We model the G-JOBASRP on a undirected graph $\mathcal{G}=(\mathcal{N}, \mathcal{A})$, where the node set $\mathcal{N}$ represents all locations in the warehouse and is denoted by $\mathcal{N} = \{0, \dots, N \}$, where $0 \in \mathcal{N}$ represents the depot. To be independent from the actual warehouse layout, we let the edge set $\mathcal{A}$ be complete and symmetric. 

We consider a set of customer orders $\mathcal{C} = \{1, \ldots, C\}$. Each customer orders $c \in \cal{C}$ consists of a number of order lines. We denote the set of all order lines by $\mathcal{I} = \{1, \ldots, I\}$, and we let $\mathcal{I}^c \subset \mathcal{I}$ be the order lines corresponding to customer request $c \in \cal{C}$. It holds that $\cup_{c \in \cal{C}} \mathcal{I}^c = \mathcal{I}$ and $\mathcal{I}^c \cap \mathcal{I}^{c'} = \emptyset$ for all $c \neq c' \in \mathcal{C}$.

Each customer order $c \in \cal{C}$ consists of either products to be returned or products to be picked. Let $q_i^r$ and $q_i^p$ denote the quantity to pick and the quantity to return for each order line $i \in \cal{I}$. The weight of the products associated with order line $i$ are denoted by $w_i$. The distance between any pair of locations $n_1, n_2 \in \mathcal{N}$ is denoted by $d_{n_1n_2}$. We let $n(i)$ denote the location in the warehouse of the product(s) associated with order line $i \in \cal{I}$.

A set of order pickers $\mathcal{E} = \{1, \ldots,, E\}$ is available for processing the customer orders. We consider a continuous time horizon of length $T$. Order pickers can perform multiple routes through the warehouse, and the order lines processed in a route will be called a batch. The set of batches of order picker $e \in \mathcal{E}$ is then given by $\mathcal{H}^e = \{1, \ldots, H_e\}$. All order pickers need to have finished their batches before the end of the time horizon. The capacity of the device of each order picker equals $Q$, and will be referred to as the order pickers' capacity. A break between the subsequent routes of an order picker is required, and equals $c_\text{break}$ minutes. Order pickers are assumed to have a constant travel time $v$, and processing an order line includes a constant pick (or return) time $s_{\textsc{pick}}$. 

Each order line $i \in \cal{I}$ has a deadline $\bar{d}_i$ before its need to be processed. Deadlines of the order lines consisting of products to be returned are set equal to $T$. If an order line is processed after the deadline, tardiness costs $\alpha$ are incurred per order line per time unit. If order lines belonging to the same customer order are assigned to multiple batches, split-up costs $\beta$ are incurred per additional split-up. For example, if three order lines are each processed in a different batch, two times $\beta$ split-up costs are incurred as the order is split up twice more than minimally needed.

We introduce decision variables $\chi^{he}_{ij}, \ \psi^{he}_{ij}$, and $\omega^{he}_{ij}$ to model the routing decisions, comparable to the formulation provided by \cite{mosheiov1994}. Thus, $\chi^{he}_{ij}$ is a binary variable that expresses whether the edge between $i, j \in N$ is traveled by picker $e \in \mathcal{E}$ in batch $h \in \mathcal{H}^e$. In turn, $\psi^{he}_{ij}$ denotes the load that has already been picked, and $\omega^{he}_{ij}$ refers to the load of order lines that still have to be returned in this route. Both loads are carried along the arc $(i,j)$; the sum must not exceed the transport capacity $Q$. The binary variable $b_{i}^{he}$ describes whether order line $i$ is contained in batch $h$ of picker $e$, and the binary variable $\tilde{b}_c^{he}$ describes whether an order line of customer $c$ is contained in batch $h$ of picker $e$.

Then the G-JOBASRP asks for solving,
\begin{align}
\min \quad  & s\cdotp\sum_{e \in \mathcal{E}} \sum_{h \in \mathcal{H}^e} \sum_{i \in \mathcal{I}} \sum_{j \in \mathcal{I}} (\frac{1}{v} d_{n(i)n(j)} + s_{\textsc{pick}}) \chi^{he}_{ij}  + \alpha \sum_{i \in \mathcal{I}} \tau_i + \beta \sum_{e\in \mathcal{E}}\sum_{h \in \mathcal{H}^e} (\tilde{b}_c^{he} - 1) &  \label{objective}
\end{align}

\vspace{-0.8cm}
\begin{align}
\text{s.t.} \quad & \sum_{e \in \mathcal{E}}\sum_{h \in \mathcal{H}^e} b_{i}^{he} = 1 & \forall i \in \mathcal{I}, \label{onebatch} \\
&  \tilde{b}_{c}^{he} \geq b_{i}^{he} & \forall (c,i) \in (\mathcal{C}, \mathcal{I}^c), h \in \mathcal{H}, e \in \mathcal{E}, \label{btilde} \\
&\sum_{e \in \mathcal{E}} \sum_{h \in \mathcal{H}^e} \sum_{i \in \mathcal{I}} \chi^{he}_{ij} = 1 & \forall j \in \mathcal{I}\backslash \{ 0 \}, \label{visitonce}\\
& \sum_{e \in \mathcal{E}}\sum_{h \in \mathcal{H}^e} \sum_{j \in \mathcal{I}} \chi^{he}_{ij} = 1 & \forall i \in \mathcal{I}
\backslash \{ 0 \}, \label{leaveonce}\\
&\sum_{j \in \mathcal{I}} \chi_{0j}^{he} \leq 1 & \forall h \in \mathcal{H}^e, e \in \mathcal{E},  \label{depotonce}\\
& \sum_{j \in \mathcal{I}} \psi^{he}_{ij} - \sum_{k \in \mathcal{I}} \psi^{he}_{ki} = \left\{\begin{matrix}
								q^p_i \cdotp w_i \cdotp b_{i}^{he} & \text{ if }	i \neq 0 \\
								-\sum_{l \in \mathcal{I}} q^p_l \cdotp w_l \cdotp b_{l}^{he} & \text{ if } i=0
								\end{matrix} \right.  & \forall h \in \mathcal{H}^e, e \in \mathcal{E}, \label{pickcontrol}\\
&\sum_{j \in \mathcal{I}} \omega^{he}_{ij} - \sum_{k \in \mathcal{I}} \omega^{he}_{ki} = \left\{ 	\begin{matrix}
								-q^d_i \cdotp w_i \cdotp b_{i}^{he} & \hspace{0.23cm} \text{ if }	i \neq 0 \\
								\sum_{l \in \mathcal{I}} q^d_l \cdotp w_l \cdotp b_{l}^{he} & \hspace{0.23cm} \text{ if } i=0
								\end{matrix} \right. & \forall h \in \mathcal{H}^e, e \in \mathcal{E},  \label{returncontrol}\\					
& \psi^{he}_{ij} + \omega^{he}_{ij} \leq Q\chi_{ij}^{he} & \forall i,j \in \mathcal{I}, h \in \mathcal{H}^e, e \in \mathcal{E}, \label{capacity}\\
& \chi^{he}_{ij} \leq b_{i}^{he} & \forall i,j \in \mathcal{I}, h \in \mathcal{H}^e, e \in \mathcal{E}, \label{chicontroli}  \\
& \chi^{he}_{ij} \leq b_{j}^{he} & \forall i,j \in \mathcal{I}, h \in \mathcal{H}^e, e \in \mathcal{E}, \label{chicontrolj}  \\
&\textsc{st}^{he} = 0  & h = 1,  \forall e \in \mathcal{E}, \label{stco1} \\
&\textsc{co}^{he} = \textsc{st}^{he} + \frac{\textsc{td}^{he}}{v} & \forall h \in \mathcal{H}^e, e \in \mathcal{E}, \label{stco2} \\ 
&st^{he} = co^{(h-1)e} + s_{\textsc{break}} & \forall h \in \mathcal{H}^e \backslash \{1\}, e \in \mathcal{E}, \label{stco3} \\
&c_i = \sum_{e \in \mathcal{E}} \sum_{h \in \mathcal{H}^e} c_i^{he} & \forall i \in \mathcal{I},  \label{ci2} \\
&\textsc{co}^{he} - c_i^{he} = \kappa^1_{ihe} & \forall i \in \mathcal{I}, h \in \mathcal{H}^e, e \in \mathcal{E}, \label{linearcihe1} \\
& c_i^{he} = \kappa^2_{ihe} & \forall i \in \mathcal{I}, h \in \mathcal{H}^e \forall e \in \mathcal{E}, \label{linearcihe2}\\
& \kappa^1_{ihe} \leq (1-b_i^{he}) M & \forall i \in \mathcal{I}, h \in \mathcal{H}^e, e \in \mathcal{E},\label{linearcihe3}\\ 
& \kappa^2_{ihe} \leq b_i^{he}M & \forall i \in \mathcal{I}, h \in \mathcal{H}^e, e \in \mathcal{E}, \label{linearcihe4}\\
&\tau_i  \geq c_i - \bar{d}_i & \forall i\in\mathcal{I}, \label{tau2} \\
&\chi^{he}_{ij}, \ b_{i}^{he}, \ \tilde{b}_c^{he} \in	\{ 0,1\} & \forall i,j \in \mathcal{I}, h \in \mathcal{H}^e, e \in \mathcal{E}, \\
& c_i, c_i^{he}, \psi^{he}_{ij}, \omega^{he}_{ij}, \kappa^1_{ihe}, \kappa^2_{ihe}, \tau_i \geq 0 & \forall i,j \in \mathcal{I}, h \in \mathcal{H}^e, e \in \mathcal{E}.
\end{align}

The Objective \eqref{objective} includes the total travel and pick time costs of all order pickers, as well as the penalty cost $\alpha$ for order lines fulfilled too late. Constraint \eqref{onebatch} ensures that each order line is contained in exactly one batch. With Constraints \eqref{btilde}, we ensure the correct modeling of $\tilde{b}_c^{he}$. With  Constraints \eqref{visitonce} - \eqref{leaveonce} we confirm that each location at which order lines need to be picked or returned gets visited exactly once in one route. Constraints \eqref{depotonce} prohibits multiple visits to the depot within one route, to avoid that the capacity restriction can be eluded. With Constraints \eqref{pickcontrol} and \eqref{returncontrol}, we keep track of the currently transported load during each tour. This is required as we have both pickups and deliveries which cause the order picker's load to fluctuate within a single tour. Constraints \eqref{capacity} limits the transported load at any point of an order picker's route to the maximum transport capacity $Q$. Constraints \eqref{chicontroli} and \eqref{chicontrolj} are required to express that an arc $(i,j)$ can only be traveled if $i$ and $j$ are contained in the same batch. In Constraints \eqref{stco1} - \eqref{stco3}, we express the relationships of the start and completion times of the batches addressed by same order picker. Constraints \eqref{ci2} define the completion time of an order line, and Constraints \eqref{linearcihe1} - \eqref{linearcihe4} are linearized constraints ensuring that the $\kappa$ variables are correct. Finally, Constraints \eqref{tau2} model the tardiness variable $\tau_i$. 


To provide insights into the problem complexity, let us first consider the order picker routing problem with integrated order picking and return handling alone. For one order picker and one batch, this problem reduces to a classical Traveling Salesman Problem (TSP), if the transport capacity of the order picker is large enough \citep{hernandez-perez2004}. Concerning TSP problems for warehouse routing, some previous research studies propose polynomial-time solution approaches for specific layouts \citep{ratliff1983, roodbergen2001}. However, \cite{theys2010} argue that the problem becomes rather intractable for warehouse layouts with more than three cross aisles. Namely, the time complexity is polynomial in the `given' number of cross ailes. The inclusion of deadlines does not complicate the decision variant of our optimization problem since deadline exceedances are only penalized. Thus, the decision variant of our optimization program combines several NP-hard routing problems with simultaneous order picking and return processing. 


\subsection{Extended MIP formulations}
We employ two different Dantzig-Wolfe reformulations on the above compact formulation (with a polynomial number of variables and constraints) in order to provide a formulation more suitable for solving large-scale instances. Both will be used in the ALNS, as described in Section \ref{sec:approach}. 

In the first formulation, we consider the exponentially large sets of order picker routes $\hat{\cal{B}}$, which we need to assign to and sequence for the multiple order pickers. Note that with a route $\hat{b} \in \cal{B}$ we denote the complete description of an order picker's route, which we need to assign to a batch $h \in \cal{H}^e$ for some order picker $e \in \cal{E}$. 

In the second formulation, we consider order picker schedules $\hat{\cal{S}}$. A schedule $s \in \hat{\mathcal{S}}$ is a completely described sequence of order picker routes that should be assigned to an available order picker. It is clear that $\hat{\cal{S}}$ is of an even larger size than $\hat{\cal{B}}$, as each schedule $\hat{s} \in \hat{\cal{S}}$ is a feasible assignment of routes $\hat{b} \in \hat{\cal{B}}$ to an order picker. 

A route $\hat{b} \in \hat{\cal{B}}$ has an associated travel time $\textsc{tt}^{\hat{b}}$. Let $x^{\hat{b}}_{eh}$ be a binary decision variable indicating whether route $\hat{b}$ is scheduled as the $h$-th batch of order picker $e$. The parameter $\xi_i^{\hat{b}}$ equals 1 if order line $i$ is processed by route $b$ and equals 0 otherwise. The first dantzig-wolfe reformulation then replaces constraints \eqref{btilde} - \eqref{chicontrolj} by the integer hull described by $\{x_{eh}^{\hat{b}} \mid  \hat{b} \text{ is a feasible route} \}$. That is, we can solve the G-JOBASRP by the following MIP formulation:

\begin{align}
\min \quad  & \sum_{e \in \mathcal{E}}\sum_{h \in \mathcal{H}^e} \sum_{\hat{b} \in \hat{\mathcal{B}}} c^{\hat{b}}x^{\hat{b}}_{eh} +  \alpha \sum_{i \in \mathcal{I}} \tau_i + \beta \sum_{i \in \mathcal{I}} \left( \sum_{e \in \mathcal{E}}\sum_{h \in \mathcal{H}^e} \sum_{\hat{b} \in \hat{\mathcal{B}}} \xi^{\hat{b}}_i x^{\hat{b}e} \ - \ 1\right)    \label{eqdw2:objective} \\
\text{s.t.} \quad & \sum_{e \in \cal{E}} \sum_{h \in \mathcal{H}^e} \sum_{ \hat{b} \in \hat{\mathcal{B}}} \xi_i^{\hat{b}} x^{\hat{b}}_{he}  \geq 1 & \forall i \in \mathcal{I} \backslash \{0\}, \label{eqdw2:all_orders} \\
& \sum_{\hat{b} \in \hat{\mathcal{B}}} x^{\hat{b}}_{he} \leq 1 & \forall h \in \mathcal{H}^e, e \in \mathcal{E}, \label{eqdw2:single_batch_per_block}\\
&\textsc{st}^{1,e} = 0  & \forall e \in \mathcal{E}, \label{eqdw2:stco1} \\
&\textsc{co}^{he} = \textsc{st}^{he} + \sum_{\hat{b} \in \hat{\mathcal{B}}} x^{\hat{b}}_{he}\textsc{tt}^{\hat{b}}  & \forall h \in \mathcal{H}^e, e \in \mathcal{E}, \label{eqdw2:stco2} \\ 
&\textsc{st}^{he} = \textsc{co}^{h-1,e} + s_{\textsc{break}} & \forall h \in \mathcal{H}^e \backslash \{1\}, e \in \mathcal{E}, \label{eqdw2:stco3} \\
&c_i = \sum_{e \in \mathcal{E}} \sum_{h \in \mathcal{H}^e} c_i^{he} & \forall i \in \mathcal{I},  \label{eqdw2:ci2} \\
&\textsc{co}^{he} - c_i^{he} = \kappa^1_{ihe} & \forall i \in \mathcal{I}, h \in \mathcal{H}^e, e \in \mathcal{E}, \label{eqdw2:linearcihe1} \\
& c_i^{he} = \kappa^2_{ihe} & \forall i \in \mathcal{I}, h \in \mathcal{H}^e \forall e \in \mathcal{E}, \label{eqdw2:linearcihe2}\\
& \kappa^1_{ihe} \leq \left(1 - \sum_{\hat{b} \in \hat{\mathcal{B}}} \xi_i^{\hat{b}} x^{\hat{b}}_{eh} \right) M & \forall i \in \mathcal{I}, h \in \mathcal{H}^e, e \in \mathcal{E},\label{eqdw2:linearcihe3}\\ 
& \kappa^2_{ihe} \leq M \sum_{\hat{b} \in \hat{\mathcal{B}}} \xi_i^{\hat{b}} x^{\hat{b}}_{eh}  & \forall i \in \mathcal{I}, h \in \mathcal{H}^e, e \in \mathcal{E}, \label{eqdw2:linearcihe4}\\
&\tau_i  \geq c_i - d_i & \forall i\in\mathcal{I}, \label{eqdw2:tau2} \\
& x^{\hat{b}}_{eh} \in	\{ 0,1\} & \forall \hat{b} \in \hat{\mathcal{B}}, e \in \mathcal{E}, h \in \mathcal{H}^e, \\
& c_i, c_i^{he}, \kappa^1_{ihe}, \kappa^2_{ihe}, \tau_i \geq 0 & \forall i,j \in \mathcal{I}, h \in \mathcal{H}^e, e \in \mathcal{E} \label{eqdw2:domain}.
\end{align}
Here, Constraints \eqref{eqdw2:all_orders} ensure that all order lines are scheduled and Constraints \eqref{eqdw2:single_batch_per_block} ensures that not more than a single route is assigned at the $h$-th batch of order picker $e$. The remaining constraints are similar as in the compact MIP formulation, i.e., they ensure the correct modeling of the start and end times of the assigned routes to the batches. 

We perform a further Dantzig-Wolfe reformulation of the above formulation by describing a sequence of routes, or schedule $\hat{s} \in \hat{\mathcal{S}}$. We let $y^{\hat{s}}$ be the binary variable equalling 1 if schedule $\hat{s} \in \hat{\mathcal{S}}$ is selected, and 0 otherwise. The parameter $\eta_c^{\hat{s}}$ indicates how many routes in $\hat{s}$ contain order lines of customer order $c$, and similarly to the previous formulation, we let $\xi^{\hat{s}}_i$ be equal to 1 if order line $i$ is contained in schedule $s$, and 0 otherwise. We denote with $\hat{c}^{\hat{s}}$ the costs of selecting schedule $\hat{s}$. These costs include the costs for tardiness and the travel costs,
\begin{align}
\min \quad & \sum_{s \in \mathcal{S}} c^{\hat{s}}y^{\hat{s}}  + \beta \sum_{c \in \mathcal{C}} \left( \sum_{\hat{s} \in \hat{\mathcal{S}} } \eta^{\hat{s}}_c y^{\hat{s}} \ - \ 1\right) & \label{eqdw1:objective} \\
\text{s.t.} \quad & \sum_{\hat{s} \in \hat{\cal{S}}} \xi_i^{\hat{s}}y^{\hat{s}} 
\geq 1 & \forall i \in \mathcal{I}\backslash \{ 0 \}  \label{eqdw1:all_jobs}\\
& \sum_{\hat{s} \in \hat{\mathcal{S}}} y^{\hat{s}} \leq K & \label{eqdw1:pickers} \\
& y^{\hat{s}} \in \{0, 1\} & \forall \hat{s} \in \hat{\mathcal{S}} \label{eqdw1:domain}
\end{align}

In the above formulation, Objective \eqref{eqdw1:objective} minimizes the costs of the selected schedules and customer order split-up costs. Constraints \eqref{eqdw1:all_jobs} ensure that all order lines are contained in the solution. With Constraints \eqref{eqdw1:pickers}, we ensure that at most $K$ schedules are selected. Finally, Constraints \eqref{eqdw1:domain} model the domain of the variables.

Both Dantzig-Wolfe reformulations presented above can, in principle, be solved with column-generation methods. Namely, we can consider subsets of the set routes and schedules and iteratively create routes and schedules that have a negative reduced cost. When no more batches and schedules of negative reduced cost can be found, we have an optimal solution of the linear relaxation of both the Dantig-Wolfe reformulations. However, considering the practically-sized instance sizes we are interested in (1000's of order lines), it is deemed impossible to solve such formulations to optimality. 

However, we will employ both formulations in the Adaptive Large Neighborhood Search, as presented in Section \ref{sec:approach}. Namely, we will store in memory all the promising batches and schedules that are found during the search, and try to solve MIPs for those fixed sets of batches and schedules. Crucial is the selection of batches and schedules during the ALNS run, on which we will elaborate in Section \ref{sec:approach}.

\section{Adaptive Large Neighborhood Search}
\label{sec:approach}

Adaptive Large Neighborhoud search, first introduced by \cite{ropke2006adaptive}, consists of iterative destroying and repair parts of a solution smartly and efficiently. Ater creating an intial solution, we destroy a solution using so-called destroy operators, according to criteria inspired on the G-JOBASRP structure. The removed parts of the solution, i.e., the order lines to be scheduled and assigned to pickers, need to be reinserted. This is done by repair operators, that typically work on a greedy basis. By smartly combining destroy and repair operators, we aim to find an improved solution. 

Only destroying and repairing the solution is insufficient to reach high-quality solutions. After a single destroy and repair operation, the solution is either worsened or improved. In the case of the latter, we consider this solution. When the solution is worsened, it is determined by a simulated annealing criterium \citep{kirkpatrick1983optimization} whether or not we accept this worsened solution. In the early stages of the heuristic, it is allowed to move to (slightly) worse solutions, while towards the end it is ensured that solutions can only improve.

A traditional ALNS implementation considers destroy and repair operators independently from each other. As the G-JOBASRP is a complex problem with many facets on which improvements are possible, not all the repair and destroy operators can be combined logically with each other. We, therefore, consider fixed pairs of repair and destroy operators that focus on improving one of the many aspects of the G-JOBASRP. We call such a fixed combination of destroy and repair operators an `operator'. 

Second, the ALNS is programmed in parallel to enhance consistency of the outcomes. We employ four different threads, each starting with a different initial solution. Then, the threads run the ALNS (as will be described in the following) independently, except that every $n^2_{\textsc{iter}}$ iterations all the threads move to the best solution found so far.

During the ALNS, we allow for infeasible solutions with respect to the capacity of the pickers. We penalize capacity violations with a cost that increases over the run of the algorithm, ensuring that at the end, it is sufficiently high so that the final solution is feasible.

Finally, we adaptively select operators. We keep track of the performance of the operators by increasing their so-called score if applying the operator resulted in an improved solution. The adaptive mechanism ensures that operators that have been proven successful are selected more frequently, which is particularly useful if we have high split-up costs.

In the following, we first discuss the basis of our generic insertion heuristic. After that, we discuss the construction of initial solutions, the operators included, and the details on the overall structure. The general flow is depicted in Algorithm \ref{alg1}, which is discussed in detail at the end of this section. We give a high-level description of the heuristic and its operators; for implementation details and actual parameter values, we refer to Appendix A.

\subsection{Generic repair heuristic (CI heuristic)}
The basis of every repair operator is a classic Cheapest Insertion (CI) heuristic. Classic CI repairs a solution by iteratively inserting the order line resulting in the lowest cost-increase. As this operation is of time complexity $O(|\mathcal{I}|^3)$, it becomes somewhat limiting for large instances or a large number of order lines to be (re)inserted.

We, therefore, take a similar approach to \cite{schrotenboer2018coordinating}. Namely, we insert the order lines in a fixed sequence.  This sequence is determined depending on the goal of the operator. Examples include sorting the order lines to be inserted lexicographically on first their aisle and second their location in the aisle or sequencing them in random order. Given this sorted (on any specified criteria) sequence of order lines, inserting them one-one-by-one is of time complexity $O(|\mathcal{I}|^2)$ plus some fixed (negligible) time for generating the sorted sequence.  The pseudocode of this procedure is presented in Algorithm \ref{alg:ci}.

In addition, it is well-known \citep[see, e.g.,][]{ropke2006adaptive} that additional randomness (or penalizing of particular characteristics) during the repair of the solution is suitable for diversification of the search. Therefore, as will be detailed in the following, evaluating insertion locations for the order lines is not for each operator solely based on the actual costs in the resulting solution. Some random costs (called noise) might be included, or particular characteristics of the solution might be penalized. We indicate this clearly when detailing the operators.

In the remainder of this section, we will refer to this generic repair operator as the `CI heuristic'.

\begin{algorithm}[t]
  \footnotesize
  \SetAlgoLined
  \KwData{Solution $s$, sorting criteria $c$, insertion criteria $i$, set of order lines $U$ }
$U \leftarrow $ SortSequenceByCriteria($U$, $c$) \;

\ForEach{$v \in U$}
{
  \text{InsertOrderLineByCriteria}($s$, $v$, $i$, $op$)\;
}
  \caption{The generic cheapest insertion (CI) heuristic. }
  \label{alg:ci}
\end{algorithm}
\subsection{Initial Solution}
For each employed thread, we construct two types of starting solutions. We first sort the order lines randomly to ensure ties are broken arbitrarily. Subsequently, we sort the order lines lexicographically on their corresponding deadlines and aisles. Then, we perform the CI heuristic on this list of sorted order lines while we assure that the first batches of all order pickers are filled before the order lines are included in the second batch of any order picker. This results in the first solution.

The procedure for the second solution is equal, except that the lexicographical sort is extended by first sorting on the customer id of each order line. This is motivated by the fact that order lines associated with a customer order are less likely to be split-up amongst multiple batches. This is especially useful if the split-up costs are relatively large.

The initial solution from which the further ALNS algorithm departs is the solution with the lowest objective, for each thread. Preliminary experiments have shown that the initial solutions differ significantly between the threads, so no additional effort is undertaken to diversify the ALNS at this point.

\subsection{Operators}
\label{sec:operators}
Each iteration of the ALNS procedure consists of adaptively choosing an operator. Each operator starts with an equal probability of being selected. If an operator improves the current solution, we slightly increase its probability of being selected in subsequent moves. If the probability exceeds a certain threshold, we reset its probability to its initial value. In this way, we ensure diversity in the search

We employ 11 different operators, each of which address different attributes of the solution, which together cover improvement in each aspect of the multi-faceted objective of the G-JOBASRP. Some of the operators change focus during the ALNS procedure by adjusting their corresponding search region, while the other operators' search space is kept constant.  In the following, we discuss each operator

\subsubsection{Similarity of Batches (Operator 1)}
For any pair of batches $\hat{b}_1$ and $\hat{b}_2$, we call these batches similar if the following condition is satisfied:
\begin{align}
\Big[ |\textsc{st}_{B_1} - \textsc{st}_{B_2} | \leq \textsc{tol}_{\textsc{sim}} \land |\textsc{co}_{B_1} - \textsc{co}_{B_2}| \leq \textsc{tol}_{\textsc{sim}}\Big], 
\end{align}
where $\textsc{tol}_{\text{sim}} \geq 0$ is a tolerance parameter that is high in the beginning the ALNS and decreases during execution. Thus, any pair of batches are similar if they have similar start times and similar completions times according to the current tolerance value $\textsc{tol}_{\textsc{sim}}$.

This operator then considers, at random, each combination of two batches (ensuring each batch is selected at most once). If two batches are similar according to the definition mentioned before, we perform a so-called single linkage cluster merging of both batches. 

In single linkage cluster merging, we iteratively group the order lines of the two similar batches to arrive at two new batches. We start with the complete set of order lines where each order line defines a group by itself. We search for two groups of order lines $k_1 \in \hat{\cal{B}}, k_2 \in \hat{\cal{B}}$ that satisfies one of the following conditions (in the presented order)

\begin{align*} 
\{k_1,k_2\} & = \{k_1, \ k_2 \subseteq \mathcal{K} \mid \exists \ i_1 \in k_1, i_2 \in k_2, c \in \mathcal{C} : i_1, i_2 \in \mathcal{I}^c  \}, \text{ or }, \\
\{k_1,k_2\} & = \arg\!\min_{k_1, \ k_2 \subseteq \mathcal{K}} \left\{ d_{n(i_1),n(i_2)} | \ i_1 \in k_1, \ i_2 \in k_2, \ k_1 \neq k_2 \right\},
\end{align*}
where $\mathcal{K}$ denotes the set of all clusters, $i_1 \in k_1$ and $i_2 \in k_2$ denote an order line in group $k_1$ and $k_2$, respectively, and $n(i_1)$ and $n(i_2)$ denote the location of order line $i_1$ and $i_2$, respectively. In words, we start with each order line of the two similar batches as a cluster. Then, we iteratively merge two clusters of which the order line are the `closest' to each other. The cluster merging procedure stops if two clusters are left or if the total load of order lines or returns reaches the transport capacity. In the latter cases, all remaining clusters are merged to a single second cluster. In earlier iterations of the ALNS, this might be impossible as the capacity violation penalty is not sufficiently high yet. Possible remaining order lines are inserted using the CI heuristic.

To obtain a new solution, we specify the sequence of locations to be visited in an S-shape routing plan while respecting the capacity restriction. That is, the locations are visited in an S-shaped route through the warehouse and the orders that would exceed the capacity get skipped, to be picked on the order picker's way back to the depot. 


\subsubsection{Outlier of Order Lines (Operator 2)}
This operator focusses on removing outliers from the solution. For each batch in the current solution, we define an outlier as an order line for which long detours are  made. A long detour is quantified in the following way. We define a threshold $0 \leq \textsc{op}^2_\textsc{tol} \leq 1$, and count the number of order lines processed in any of the aisles in the current batch. If the fraction of order lines in an aisle is smaller than $\textsc{op}^2_\textsc{tol}$, all order lines in the corresponding aisle are marked as outliers.

The operator then searches randomly through all order lines in the current solution.  At most $\textsc{op}^2_\textsc{int} * N$ order lines are checked on being an outlier and subsequently removed if it qualifies as an outlier. Here, $0 \leq \textsc{op}^2_\textsc{int} \leq 1$. The insertion is done using the CI heuristic. Here, the sorting criterium is based on the associated aisle of the order lines to be reinserted, and the insertion criterium is the actual insertion costs in the resulting solution.

\subsubsection{Order line destroy and repair (Operator 3)}
Operators 3 is based on classical ALNS moves. First, the operator selects a random number of order lines to remove from the current solutions, uniformly distributed between $\textsc{op}^{3a}_\textsc{int}\cdot N $ and $\textsc{op}^{3b}_\textsc{int} \cdot N$. The removed order lines are then placed in a random order, and inserted using the CI heuristic.

\subsubsection{Batch destroy and repair (Operator 4)}
Operator 4 works in a similar fashion as Operator 3, only here we remove complete batches from a solution. The number of removed batches is uniformly drawn between 1 and $\textsc{op}^{4}_\textsc{int} \cdot E \cdot H$. All the order lines associated with the removed batches are placed in a random order, and inserted using the CI heuristic.

\subsubsection{Aisle destroy and repair (Operator 5)}
This classical ALNS move aims to improve the solution by focussing on order lines that are located in the same aisle. We select a random number of batches, uniformly drawn between 1 and $\textsc{op}^5_{\textsc{int}} \cdot E \cdot H$. We insert the customers based on the CI heuristic. As a sorting criterium, we use the aisle associated with the order line (ties are broken randomly). 

\subsubsection{Customer order destroy and repair (Operator 6)}
Operator 5 removes a uniformly randomly drawn number between $\textsc{op}^{6a}_\textsc{int}\cdot C $ and $\textsc{op}^{6b}_\textsc{int} \cdot C$ complete customer orders from the current solution. We then randomly sort the customer orders (with associated order lines grouped), and insert them using the CI heuristic. We ensure to insert all the order lines associated with the same customer order in a single batch. We do not allow for capacity violations while inserting the customer orders, which can lead to an unrepaired final solution. If this is the case, we insert the remaining customersusing the CI heuristic in the complete solution without grouping together the order lines associated with the same customer order.

\subsubsection{Split customer order destroy and repair (Operator 7)}
This operator is identical to Operator 5, but only removes the associated order lines of customer orders that are split among multiple batches. The maximum of such removed customer orders from the current solution is uniformly drawn between $\textsc{op}^{7a}_\textsc{int}\cdot C $ and $\textsc{op}^{7b}_\textsc{int} \cdot C$.

\subsubsection{Batch variable neighborhood decent (Operator 8)}
Operator 8 focusses solely on improving the order picker routes. It employes the relocate, swap, and 2-opt local search operators part of the genetic algorithm by \cite{schrotenboer2017order} in a variable neighborhood decent way. First, relocate moves are applied until no improvement is found. Then, swap moves are applied until no improvement is found. Finally, 2-opt moves are applied until no improvement is found. If either swap or 2-opt finds an improvement, we restart from the beginning. The operator terminates if all operators are unable to find an improvement. We refer to \cite{schrotenboer2017order} for the details on the implementation of the relocate, swap, and 2-opt local search operators.

\subsubsection{Tardines (Operator 9)}
The Tardiness operator focusses on improving the solution concerning the current tardiness of the solution. Here, tardiness equals the absolute difference between the time the order line is dropped at the depot and its deadline. We randomly check each order line (with a maximum of $\textsc{op}^{9}_\textsc{int}\cdot N $) for its so-called tardiness. We remove the order line from the solution if the deadline is exceeded or if it is dropped at the depot at least $\textsc{op}^{9}_\textsc{tol}$ before its deadline.

We then sort the orders based on their deadline and insert the sorted sequence of customers via the CI heuristic. We additionally penalize insertion in the solution based on the difference in ready-time and the orders corresponding deadline. Namely, if the deadline is respected, we add $\textsc{op}^{9}_\textsc{pen}$ times the difference between depot drop time and the deadline to the actual insertion costs. If the deadline is exceeded, we add a sufficiently big penalty to prefer any insertion that respects the deadline. In this way, both an early and a late pick are penalized. Preliminary experiments have shown that penalizing early picks is helpfull for a faster convergence of the ALNS.

\subsubsection{Random customer removal, sorted insert (Operator 10)}
This operator works similarly to Operator 6. We remove a random number of customer, drawn between the same bounds as described for Operator 6. Then, instead of assuring that the customers can be inserted in the same batch, we use as a sorting criterium for the CI heuristic the customer index associated to the order lines. 

\subsubsection{Random batch removal, sorted insert (Operator 11)}
This operator is a blend of Operator 10 and Operator 2. We first select several batches uniformly within the same bounds as indicated for Operator 2. Then, we sort the removed order lines based on their customer index and apply our CI insertion heuristic. 

\subsection{MIP-based improvements}
Next to the metaheuristic type of operators as discussed in Section~4.2, we employ two different Mixed Integer Programming based improvement operators based on the Dantzig-Wolfe decomposed models presented in Section~3.

The first MIP operator solely focusses on an optimal assignment of the batches currently part of the solution. It is simply obtained by solving the extended formulation \eqref{eqdw2:objective} - \eqref{eqdw2:domain} with the set of batches $\hat{\mathcal{B}}$ equal to the batches currently in the solution. 

The second MIP operator uses information of the complete ALNS run so far. While executing the ALNS, many different solutions with different batches are encountered. We store all the unique batches  in the set of batches $\hat{\mathcal{B}}$. For this set of batches, we can solve the extended formulation of the G-JOBASRP. However, preliminary experiments have shown that solving this for the complete set of batches is computationally expensive. We, therefore, select an arbitrary number of batches $\textsc{op}^{\textsc{mip}}_{\textsc{int1}} \cdot E \cdot H$ which we fix in the current schedule. We then solve the extended formulation, which will then assign batches (and start and end times) to the non-fixed batches. This is repeated $\textsc{op}^{\textsc{mip}}_{\textsc{int2}} \cdot E \cdot H$ times. 

We perform these MIP based operators in the order presented above. If an operator improves the solution we terminate the execution of all MIP-based operators.

Finally, as will be detailed in Section~\ref{sec:numericalresults}, we will use the extended MIP formulations presented in Section~3 to validate the performance of the ALNS. That is, we store all batches and schedules found during execution of the ALNS, and solve the complete extended MIP formulations based on batches and schedules.

\begin{algorithm}[t]
\footnotesize
\SetAlgoLined
$\mathbf{p} \leftarrow$ initalizeAdaptiveProb() \;
$s_{\text{cur}},s_{\text{best}} \leftarrow$ initialSolution() \;
$n_1, n_2, it \leftarrow 0$ \;
\While{\upshape $n_1 < N^1_{\text{iter}}$}
{ 
    
	\While{\upshape $n_2 < N^2_{\text{iter}}$}
    {
    $s_\text{cur} \leftarrow s_\text{best}$\;
    applyRandomOperator($\mathbf{p}$, $s_\text{cur}$)\;
    	updateProb($s_{\text{cur}}$, $s_\text{best}$, $\mathbf{p}$)\;
    	\If {\upshape \text{acceptSolution}($s_{\text{cur}} $, $s_{\text{best}}$, $it$)}
    	{
    		$s_{\text{best}} \leftarrow s_{\text{cur}} $\;
    	}
    	
    	$n_2 \leftarrow n_2 + 1$\;
    	$it \leftarrow it + 1$.
    	
    }
   
    $n_1 \leftarrow n_1 + 1$\;
    ExchangeInformationBetweenThreads($s_{\text{best}}, \mathbf{p}$)\;
    MipBasedImprovements()\;
}
 \caption{General flow of the ALNS}
 \label{alg1}
\end{algorithm}

\subsection{Overal heuristic structure}
The complete ALNS Procedure is outlined Algorithm \ref{alg1}. On lines 1-3, we initialize a vector of adaptive probabilities $\mathbf{p}$, we compute the initial solutions by calling the procedure ``initialSolution()", and we initalize some iterators to zero. Four procedures are detailed in Algorithm \ref{alg1}. First, the ``applyRandomOperator($\mathbf{p}$, $s_\text{cur}$)" selects an operator based on the probability vector $\mathbf{p}$ and applies that operator on the solution $s_\text{cur}$. Second, the ``updateProb($s_\text{cur}, s_\text{best}, \mathbf{p}$)" updates the probability vector based on the result of the operator being applied. The ``acceptSolution($s_\text{cur}, s_\text{best}, it)$" procedure contains the simulated annealing criterion (detailed below) that either accepts or denies a new solution. Fourth, the ``exchangeInformationBetweenThreads($s_\text{best}, \mathbf{p}$)" procedure exchanges information between the running threads, i.e., it updates the best solution and the adaptive probability vector. Notice that this function is called only if all the threads arrive at it simultaneously. 

As denoted in Algorithm \ref{alg1}, a simulated annealing criterion surrounds the ALNS iterations. Let $it$ be the current iteration, and let $N_{\text{iter}}^{\max} := N_\text{iter}^1N_\text{iter}^2$. Moreover, let $z_\text{best}$ be the objective value before applying an operator and let $z_\text{cur}$ be the objective value after applying. If $z^* < z$, we accept the new solution. Otherwise, we accept it with probability
$$
\exp\left(-1\frac{(z_\text{cur} - z_\text{best})/z_\text{best} \cdot \alpha }{[1 - (i/N_\text{iter}^{\max})]^\beta}\right).
$$
Here, $\alpha$ denotes how much objective values may become worse in order to still are likely to be accepted, and $\beta$ denotes how fast we aim to let the ALNS converge. 
Moreover,  $N_\text{iter}^1 \times N_\text{iter}^2$ number of iterations are being run in the ALNS. If $N_\text{iter}^2$ iterations have passed, the running threads exchange information: First, each thread continues with the solution that has the lowest objective. Second, the adaptive probabilities are updated by averaging the probabilities of each thread with the probabilities of the thread that provided the best solution.

\label{sec:iteration}

\section{Numerical Results}
\label{sec:numericalresults}
In this section, we will provide insights into two main points. First, we investigate the cost-savings potential of integrating the restocking of (returned) products in the regular order picking process. Second, we investigate the efficiency of allowing the split-up of customer orders among multiple batches. We do this by solving the G-JOBASRP on a carefully constructed testbed of instances. 

In the following, we first detail the construction of the benchmark instances used. We then comment on our preliminary computational campaign to find suitable control parameters of the ALNS. We then solve the new G-JOBASRP instances and provide some comparison with common heuristics (for example assuming fixed routing policies). We then focus on providing insights on the two main points as mentioned before.

 
\subsection{Benchmark data sets}
We perform our experiments in a rectangular warehouse layout that is commonly used to study order picking performance \citep{gademann2001, gong2009, schrotenboer2017order}. The parallel aisles have length $a_{\text{length}}$ and width $a_{\text{width}}$, along with two cross aisles at the beginning and the end of the parallel aisles. Products are stored only in the parallel aisles, not in the cross aisles. We assume that order pickers can reach locations on the left and the right side within an aisle without traveling additional distances. The distance between any two points in the warehouse is then trivially calculated as the warehouse distance between those points. 



Unfortunately, the benchmark instances of \cite{scholz2017order} are not available anymore so we could not test the performance of the ALNS on those instances. We therefore, carefully constructed  $12$ generated sample data sets, that serve as the order and return arrival process of a single working shift of $8$ hours for multiple order pickers. 

The data sets vary in the number of order lines, the warehouse size, and the number of products to be restocked, as is detailed in Table \ref{datasets}. In column $4$ in Table \ref{datasets}, we denote the percentage of customer returns with respect to the total number of requests. We ensure that a return amount of $30\%$ indicates that $30$ of $100$ customer requests are product returns, which corresponds to a $(30/70=) 43\%$ actual return rate. Pickers move through the warehouse with a constant pace of $0.7$ meters per second, and the time to pick or return one order line equals $8$ seconds. In all experiments, travel costs equal $0.54$ EUR per minute per order picker, which equates to approximately $32$ EUR for an order picker in each hour. Delay costs equal $0.06$ EUR per minute per delayed product \citep[similar to][]{tsai2008}. The capacity of the picking device is $80$ kg for all order pickers. Finally, we set the break between two subsequent routes of the same order picker to equal $c_{break} = 5$ minutes.
\begin{table}

\caption{Sample Data Sets}
\centering
\begin{tabular}{rrrrrrrrr} 
\hline
Dataset & Customers & Order lines & Return amount &  Total weight (kg) & Orders (kg) & Returns (kg) \\
\hline
1 & 195	 	& 598   & 10\% & 641.0	& 578.1	  	& 62.9	 	\\
2 & 227	 	& 640	& 20\% & 714.0 	& 579.4	  	& 134.6	 	\\
3 & 292	 	& 864	& 30\% & 904.8  & 528.6	  	& 376.2	 	\\

4 &	566 	& 1737 	& 10\% & 1893.3	& 1764.2	& 129.1		\\
5 &	672 	& 2028 	& 20\% & 2191.9	& 1766.0	& 425.9		\\
6 &	920 	& 2786 	& 30\% & 2988.8	& 2097.9	& 890.9		\\	
													
7 &	1348 	& 4000 	& 10\% & 4299.8	& 3848.7	& 451.1		\\
8 &	1706 	& 5049 	& 20\% & 5647.1 & 4597.3	& 1049.8	\\
9 &	1817 	& 5431 	& 30\% & 6092.8 & 4099.4	& 1993.4	\\[0.2cm]					
10&	2029 	& 6059 	& 10\% & 6711.6	& 5947.0	& 764.6		\\
11&	2341 	& 6995 	& 20\% & 7654.3 & 6082.5	& 1571.8	\\
12&	2673 	& 8038 	& 30\% & 8897.9	& 6181.9	& 2716.0	\\
\hline
\end{tabular}
\label{datasets}
\end{table}

The number of parallel aisles in data sets $1-9$ is $35$, with an aisle length of $30$ meters. For data sets $10-12$, we considered $50$ aisles, with a length of $40$ meters. The aisle width equals $2.5$ meters in all experiments. Deadlines for returned products are the end of the work shift (i.e., $8$ hours). Deadlines for customer orders might occur during the shift, at every full hour; we assigned them randomly with a uniform distribution. The number of products in an order line can be a maximum of $4$, being $1$ for $40\%$ of the order lines, $2$ for $30\%$, $3$ for $20\%$, and $4$ for $10\%$, resulting in a mean quantity of two products per order line. The weight of a product is accurate to $0.1$ kg with a uniform distribution and a maximum of $1$ kg. Similarly, the storage locations of order lines, given by the aisle number and the location in the aisle, accurate to $0.1$ meters, are assigned with a uniform distribution. We assumed random storage assignment as it performs well when the number of order lines to be fulfilled is large \citep{chan2011}. Finally, the datasets cover product return rates that are typical for e-commerce businesses \citep{mostard2005, koster2002}.

\subsection{Parameter tuning}
The performance of the ALNS appears rather robust concerning the set paramaters. We, therefore, briefly describe the parameters used for the results described in the following sections.

We allow $N^1_{iter} = 50$ outer iterations of $N^2_{iter} = 100$ inner iterations, see Algorithm 2. Note that this is predominantly oriented so that the largest instances still have reasonable computation times. For instance, they can be solved in a single night before operations start in the warehouse in the morning. The $\textsc{tol}_{\textsc{sim}}$ equals 2000 at the start, and is lowered linearly to 500 at the end of the algorithm. An order line is called an outlier if it is part of an aisle that is used less then 20 \% (i.e. $\textsc{op}_{\textsc{tol}}^2 = 0.2$) in a route. Consequently, $\textsc{op}_{\textsc{int}}^2$ equals 0.15, indicating that at most 15\% of the solution is destroyed. 

Then, Operator 3 removes a random number of order lines with $\textsc{op}_{\textsc{int}}^{3a} = 0.05$ and $\textsc{op}_{int}^{3b} = 0.15$. For Operators 4 and 5 (that remove batches and aisles, respectively), $\textsc{op}_{\textsc{int}}^4 = 0.25$ and $\textsc{op}_{\textsc{int}}^5 = 0.15$, respectively. 

Furthermore, Operator 6 removes a randomly number of customers uniformly according to $\textsc{op}_{\textsc{int}}^{6a} = 0.05$ and $\textsc{op}_{\textsc{int}}^{6b} = 0.15$. For Operator 7, this is 0.05 for $\textsc{op}_{\textsc{int}}^{7a}$ and 0.10 for $\textsc{op}_{\textsc{int}}^{7b}$. 

Tardiness Operator 9 checks at most $\textsc{op}_{\textsc{int}}^{9} = 0.15$ of the order lines for its tardiness contribution. We set $\textsc{op}_{\textsc{tol}}$ = 1000 and $\textsc{op}_{\textsc{pen}}^9 = 0.01$. Then, Operator 10 removes as many customers as Operator 6, and, finally, Operator 11 removes at most the fraction of $\textsc{op}_{\textsc{int}} = 0.25$ of the batches.

To imitate the probabilities guiding the adaptive aspect of the ALNS, we initialize each operator with a `score' equal to 1. The distribution of this scores among the operators then guides the operator selection. If a succesfull move is made, we increase the score of the associated operator and the 4 previous operators that resulted into accepted moves with 0.25. If after the inner (and parallel) part of the algorithm a score is larger than 5, we reset it to 1 so as to keep diversity in the choosen operators. 

Finally, the results presented in this section are obtained without using the MIP operators. We discuss the (dis)adventages of using these operators in Subsection \ref{mipop}.

\begin{table}[h!]
\def\arraystretch{0.67}

\centering
\caption{Results of comparison ALNS with BM1 and BM2 with $\beta = 0$.}
\label{tab:comparison}
\begin{tabular}{llrrrr} \toprule 
Dataset & $E$ & BM1 & BM2 & ALNS & Dif(\%) \\ \midrule
1    & 3  & 102.7 & 98.9  & 66.0  & -33.3 \\
     & 4  & 102.7 & 98.5  & 65.9  & -33.1 \\
     & 5  & 102.6 & 98.1  & 66.0  & -32.7 \\
2    & 3  & 105.6 & 102.4 & 70.4  & -31.2 \\
     & 4  & 106.3 & 102.8 & 69.7  & -32.2 \\
     & 5  & 106.3 & 102.6 & 69.7  & -32.0 \\
3    & 4  & 120.4 & 117.9 & 84.5  & -28.4 \\
     & 5  & 120.3 & 117.9 & 84.2  & -28.5 \\
     & 8  & 120.3 & 117.8 & 84.1  & -28.7 \\
4    & 7  & 229.2 & 223.2 & 175.6 & -21.3 \\
     & 8  & 230.2 & 224.0 & 173.2 & -22.7 \\
     & 9  & 229.1 & 223.8 & 173.1 & -22.7 \\
5    & 8  & 250.0 & 244.5 & 195.6 & -20.0 \\
     & 9  & 250.4 & 244.4 & 195.9 & -19.9 \\
     & 10 & 250.4 & 244.7 & 194.5 & -20.5 \\
6    & 8  & 312.4 & 308.6 & 267.3 & -13.4 \\
     & 9  & 312.7 & 308.7 & 264.6 & -14.3 \\
     & 10 & 313.4 & 308.5 & 264.0 & -14.4 \\
7    & 12 & 443.0 & 437.8 & 387.4 & -11.5 \\
     & 13 & 441.1 & 436.1 & 390.1 & -10.5 \\
     & 14 & 440.4 & 435.7 & 386.9 & -11.2 \\
8    & 14 & 538.0 & 531.3 & 481.7 & -9.3  \\
     & 15 & 536.2 & 531.3 & 481.9 & -9.3  \\
     & 16 & 537.0 & 531.8 & 482.2 & -9.3  \\
9    & 14 & 549.5 & 543.7 & 498.2 & -8.4  \\
     & 15 & 550.2 & 544.7 & 499.0 & -8.4  \\
     & 16 & 549.3 & 543.6 & 501.3 & -7.8  \\
10   & 20 & 751.5 & 742.7 & 661.0 & -11.0 \\
     & 21 & 749.2 & 739.9 & 653.6 & -11.7 \\
     & 22 & 751.5 & 742.3 & 658.1 & -11.3 \\
11   & 20 & 823.2 & 813.2 & 725.2 & -10.8 \\
     & 21 & 823.2 & 812.8 & 730.9 & -10.1 \\
     & 22 & 822.4 & 812.6 & 732.0 & -9.9  \\
12   & 20 & 899.2 & 889.7 & 815.3 & -8.4  \\
     & 21 & 901.0 & 889.2 & 810.0 & -8.9  \\
     & 22 & 898.0 & 889.6 & 813.7 & -8.5  \\ \midrule
Avg. &    & 426.9 & 421.0 & 368.7 & -17.4 \\ \bottomrule
\end{tabular}
\end{table}

\subsection{Computation times}
To not disturb the reader from the essence of our contribution, namely to show the dynamics of the G-JOBASRP, we do not accompany the detailed results with the computation times of the ALNS. We will, however, give a brief description of the computational aspects in the following.

The instance sizes range from 598 order lines to 8038 order lines. This is significantly larger compared to existing problems in related settings, as for instance, the newest proposed datasets for the (much more stylized and less constrained) capacitated vehicle routing problem range from 100 to 1000 customer \citep{uchoa2017new}. As stated before, we callibrated our parameters so that the largest instances are still solvable in reasonable time, i.e., they can be solved overnight. The ALNS is programmed in C++ and all experiments are run on a Xeon E5 2680v3 2.5 GHz CPU.

Regarding the smaller instances, the computation times range from 313 seconds for Dataset 1 to 1.5 hours for Dataset 6. Datasets 7, 8 and 9 then respectively ask approximately 4.5, 7, and 8 hours for solving the associated G-JOBASPR. Finally, Dataset 12 took around 11.5 hours to solve, which we deemed feasible for practical purposes. If this is, due to the practical circumstances, deemed infeasible, one could reduce the number of iterations to achieve only slightly worse solutions in shorter computation times.

\subsection{Benchmark models}
In practice, instances of the G-JOBASRP of the size we propose (i.e., up to 8038 order lines) are solved with intuitive constructive heuristics. We, therefore, propose two of such methods to compare the performance of ALNS. 

The first benchmark model (BM1) uses a variation of the earliest deadline first (EDF) job scheduling heuristic and organizes the route of each batch using cheapest feasible insertion. In contrast with job scheduling problems, the G-JOBASRP considers no single jobs (i.e., a batch equals  `a job' in scheduling terminology). Thus, when applying EDF, we need to make an additional decision when an order line is included in an already existing (non-empty) batch or in a new empty batch. This decision involves the consideration of the earliest deadline that occurs in the non-empty batch among already included order lines. Therefore, we consider the order lines to be sorted lexicographically according to first their deadlines and second their aisle. The first order line is included in the first batch of the first order picker. This batch is filled up with the next order lines. We determine a preliminary route by inserting the order lines using cheapest feasible insertion in the sequence in which the order lines are included in the batch. 

As soon as an order line is included in a particular batch, we also allow insertion in the same batch of the next order picker. If no such batch is available, we allow insertion of the order line in a new batch of the first order picker. 

The second benchmark model (BM2) improves upon BM1 by afterward performing a local search on the created batches. All ties while sorting the sequence of order lines are broken arbitrarily. We, therefore, perform both benchmark models 100 times and select the solution with the lowest objective.

\begin{table}[h!]
\centering
\caption{Results of solving the G-JOBASRP benchmark instances}
\setlength\tabcolsep{4pt} 
\def\arraystretch{0.67}
\label{tab:allsolution}
\begin{tabular}{lllllllllllll} \toprule
 &          &            \multicolumn{4}{c}{$\beta = 0$ (split-ups)}   &   \multicolumn{5}{c}{$\beta = 10000$ (no split-ups)}        & \multicolumn{2}{c}{dif.}  \\ \cmidrule(r){3-6} \cmidrule(l){7-11}  \cmidrule(l){12-13}
Dataset & $E$          & int.  & orders & returns                     & dif.    & int.(1) & int.(2) & orders & returns & dif.    & int. & sep.\\ \midrule
1  & 3        & 66.0  & 61.6  & 12.4  & -10.8 & 106.7  & 107.5  & 103.4  & 12.4  & -7.2  & -38.6 & -36.1 \\
   & 4        & 65.9  & 61.8  & 12.4  & -11.2 & 107.8  & 107.5  & 103.2  & 12.4  & -7.0  & -38.7 & -35.8 \\
   & 5        & 66.0  & 61.8  & 12.4  & -11.2 & 107.0  & 107.5  & 103.1  & 12.4  & -7.0  & -38.6 & -35.7 \\
2  & 3        & 70.4  & 60.0  & 24.0  & -16.1 & 108.3  & 107.5  & 96.5   & 24.0  & -10.8 & -34.4 & -30.3 \\
   & 4        & 69.7  & 59.9  & 24.1  & -17.0 & 108.0  & 107.1  & 95.9   & 24.0  & -10.6 & -34.9 & -29.9 \\
   & 5        & 69.7  & 59.7  & 23.9  & -16.6 & 106.2  & 106.4  & 94.9   & 24.0  & -10.5 & -34.5 & -29.7 \\
3  & 4        & 84.5  & 60.5  & 43.7  & -19.0 & 126.5  & 126.8  & 96.8   & 43.9  & -9.9  & -33.4 & -25.9 \\
   & 5        & 84.2  & 60.5  & 43.7  & -19.1 & 127.0  & 124.5  & 95.8   & 43.8  & -10.8 & -32.4 & -25.4 \\
   & 8        & 84.1  & 60.3  & 43.6  & -19.1 & 123.0  & 123.7  & 96.7   & 43.6  & -11.9 & -32.0 & -26.0 \\
4  & 7        & 175.6 & 167.0 & 23.6  & -7.9  & 335.2  & 329.7  & 322.3  & 23.6  & -4.7  & -46.7 & -44.9 \\
   & 8        & 173.2 & 163.0 & 23.7  & -7.3  & 332.3  & 331.8  & 318.2  & 23.7  & -2.9  & -47.8 & -45.4 \\
   & 9        & 173.1 & 163.1 & 23.7  & -7.3  & 331.7  & 334.6  & 320.5  & 23.8  & -2.8  & -48.3 & -45.7 \\
5  & 8        & 195.6 & 165.7 & 48.6  & -8.7  & 366.0  & 368.3  & 331.1  & 48.7  & -3.0  & -46.9 & -43.6 \\
   & 9        & 195.9 & 166.6 & 48.7  & -9.0  & 366.6  & 368.2  & 327.2  & 48.5  & -2.0  & -46.8 & -42.7 \\
   & 10       & 194.5 & 165.1 & 48.5  & -9.0  & 371.2  & 362.3  & 329.3  & 48.6  & -4.1  & -46.3 & -43.5 \\
6  & 8        & 267.3 & 199.8 & 89.2  & -7.5  & 470.3  & 462.4  & 396.0  & 89.8  & -4.8  & -42.2 & -40.5 \\
   & 9        & 264.6 & 198.1 & 89.1  & -7.9  & 463.7  & 460.6  & 390.2  & 89.9  & -4.0  & -42.6 & -40.2 \\
   & 10       & 264.0 & 198.1 & 89.7  & -8.3  & 466.7  & 464.4  & 391.5  & 89.1  & -3.4  & -43.1 & -40.1 \\
7  & 12       & 387.4 & 359.8 & 50.7  & -5.6  & 740.5  & 764.0  & 740.5  & 50.7   & -3.4   & -49.3 & -48.1 \\
   & 13       & 390.1 & 359.0 & 50.7  & -4.8  & 764.2  & 765.8  & 730.5  & 50.8  & -2.0  & -49.1 & -47.5 \\
   & 14       & 386.9 & 358.2 & 50.4  & -5.3  & 770.2  & 767.8  & 736.5  & 50.9  & -2.5  & -49.6 & -48.1 \\
8  & 14       & 481.7 & 413.4 & 99.9  & -6.1  & 946.0  & 925.5  & 857.0  & 100.0 & -3.3  & -48.0 & -46.4 \\
   & 15       & 481.9 & 413.2 & 99.4  & -6.0  & 932.8  & 928.7  & 857.3  & 99.3  & -2.9  & -48.1 & -46.4 \\
   & 16       & 482.2 & 413.7 & 99.9  & -6.1  & 943.1  & 936.6  & 854.4  & 100.3 & -1.9  & -48.5 & -46.2 \\
9  & 14       & 498.2 & 372.0 & 175.5 & -9.0  & 906.6  & 909.9  & 754.9  & 179.3 & -2.6  & -45.2 & -41.4 \\
   & 15       & 499.0 & 372.1 & 175.6 & -8.9  & 914.6  & 901.1  & 747.5  & 179.7 & -2.8  & -44.6 & -40.9 \\
   & 16       & 501.3 & 371.1 & 175.2 & -8.2  & 903.4  & 906.8  & 751.0  & 178.1 & -2.4  & -44.7 & -41.2 \\
10 & 20       & 661.0 & 609.6 & 91.1  & -5.7  & 1773.2 & 1606.5 & 1681.6 & 91.6  & -9.4  & -58.9 & -60.5 \\
   & 21       & 653.6 & 608.0 & 90.9  & -6.5  & 1666.3 & 1582.5 & 1575.1 & 91.1  & -5.0  & -58.7 & -58.1 \\
   & 22       & 658.1 & 610.4 & 91.1  & -6.2  & 1644.1 & 1553.1 & 1553.2 & 90.9  & -5.5  & -57.6 & -57.3 \\
11 & 20       & 725.2 & 623.8 & 168.1 & -8.4  & 1847.0 & 1676.1 & 1674.7 & 172.2 & -9.3  & -56.7 & -57.1 \\
   & 21       & 730.9 & 621.1 & 168.3 & -7.4  & 1783.2 & 1669.9 & 1625.3 & 171.8 & -7.1  & -56.2 & -56.1 \\
   & 22       & 732.0 & 624.8 & 167.7 & -7.6  & 1724.9 & 1638.7 & 1585.8 & 171.7 & -6.8  & -55.3 & -54.9 \\
12 & 20       & 815.3 & 626.2 & 275.6 & -9.6  & 1948.2 & 1941.6 & 1658.3 & 290.0 & -0.3  & -58.0 & -53.7 \\
   & 21       & 810.0 & 627.1 & 276.8 & -10.4 & 1889.7 & 1847.3 & 1599.0 & 290.8 & -2.2  & -56.2 & -52.2 \\
   & 22       & 813.7 & 626.9 & 276.6 & -9.9  & 1850.3 & 1819.6 & 1564.5 & 285.8 & -1.7  & -55.3 & -51.2 \\ \midrule
 Avg. &  &       &       &       & -9.6  &        &        &        &       & -5.1  & -46.3 & -43.5        \\ \bottomrule
\end{tabular}
\end{table}

\subsection{Solutions to the G-JOBASPR instances}
We solve the constructed datasets 1-12 with the ALNS as described in Section \ref{sec:approach}. We create three G-JOBASRP instances from each dataset, each having a different number of order pickers. These numbers and the maximum number of 8 batches for each order picker resulted from a preliminary analysis of the instances. 

We first compare the performance of the ALNS with the performance of BM1 and BM2, when $\beta = 0$. The results are provided in Table \ref{tab:comparison}. A few observations stand out. First, the average cost reduction from using ALNS instead of BM2 equals 17.4\%. Second, the instances with a higher fraction of product returns show slightly smaller cost reductions. This decrease is due to dedicated routes with only deliveries becoming efficient for an increasing number of deliveries. Third, the cost-reduction that results from using the ALNS decreases for larger instances. This decrease is, again, due to the structure of the batches for the final solutions. If there are many order lines to pick, the average number of aisles an order picker traverses is rather small. Due to sorting on aisle number and customer deadlines in BM1 and BM2, the structure of the proposed solutions of both benchmark models is rather similar. Hence, for the scenarios where the warehouse is working at its full capacity (e.g., just before Christmas), using an intuitive heuristic already performs quite well.  However, warehouses are typically not working at their full capacity. Then, using the ALNS instead of the intuitive heuristics will improve efficiency significantly (20-30\% cost reductions). 

In Table \ref{tab:allsolution}, we present the results for solving the instances associated with the datasets for $\beta= 0$ and $\beta = 10000$ with the ALNS. Here, the former indicates a situation where customer orders might be split-up without any costs, whereas the latter indicates a situation in which customer orders cannot be split. All the presented results are the minimum values over ten repetitions. The columns headed by `int.' denote the costs associated with solving the problem with the product returns integrated with the order picking process. For the case with $\beta = 10000$, we provide two integrated objectives. The costs `int.(1)'  are associated with solving the G-JOBASRP instances with the number of order pickers increased for the integrated problem by the fraction of product returns in the associated instance. The costs `int.(2)' denote the minimum of 1)solving the G-JOBASRP with the specified number of order pickers and of 2) solving the G-JOBASRP completely separately. The motivation for both cost measures is due to the observation that tardiness costs are included in the larger instances if $\beta = 10000$. In such a case, incorporating a delivery (i.e., an order line of product returns) results in longer travel times and hence more tardiness costs due to its associated product drop time. Hence, for the larger instances (datasets 10-12), it is particularly difficult to find the benefit of the integrated solution (if there is any). Moreover, as we give the same number of iterations to both settings, and the integrated setting consists of significantly larger instances, it also provides a more fair comparison in terms of the quality of the ALNS. In the final two columns, we measure the difference between $\beta=0$ and $\beta=10000$, where column `int.' denotes the differences between the integrated problems (int.(2) for $\beta=10000$) and column `sep'. denotes the difference between the separated problems.

The columns headed by `orders' and `returns' denote the costs of solving only the order picking problem and the product return problem, respectively. In other words,  we solve the G-JOBASRP separately for the pickups and the deliveries. The differences between the integrated and separated problem are denoted (in \%)  in the column `dif'.

From the results in Table \ref{tab:allsolution}, it is observed that splitting up customer orders results into an average cost-saving of 47.7\% when product returns are integrated and 43.7 \% when product returns are handled separately. Hence, the opportunity to improve efficiency by separating customer orders is very likely for practical situations. This will be evaluated further in Section \ref{sec:sens_split}.
The effect of integrating product returns leads to average cost-savings of 9.2 \% and 1.9\% when order split-ups are not penalized ($\beta = 0$) and are prohibited ($\beta = 10000$), respectively. The former cost-savings are in line with the results for the single picker, single tour scenario as studied by \cite{schrotenboer2017order}, whereas the latter cost-savings seem to be marginalized. The performance on the larger instances is driving this marginalization since it is impossible to process all customer orders before their associated deadlines in these instances.

\subsection{The effect of multiple threads}
\begin{figure}[t]
\centering
\includegraphics[scale=0.75]{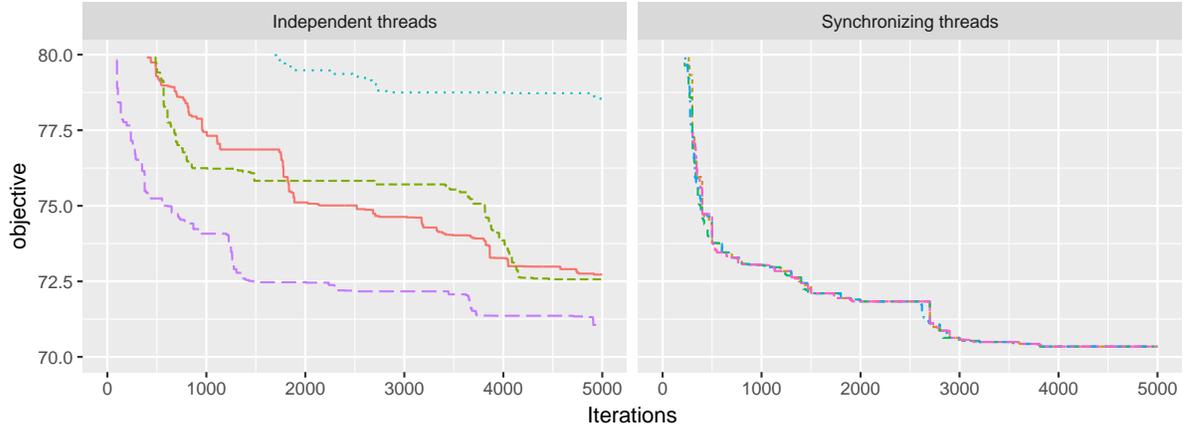}
\caption{Solution trajectory of the ALNS on dataset 2 with 3 order pickers with four threads exchanging information (on the left) and four threads working indepedently (on the right). Each line in each figure corresponds to the current best-known solution found by a single thread.} 
\label{fig:par1}
\end{figure}

As explained in Section \ref{sec:approach}, the ALNS is programmed in parallel using four threads. A simple, though computationally useful feature is that we exchange information between the threads every 100 iterations. In this section, we briefly illustrate the impact of this on the observed solution trajectory.

We compare the ALNS with threads exchanging information and the best-known solution with the same ALNS but now run by four threads on isolation. We focus on Dataset 2 with three order pickers and Dataset 6 with eight order pickers. Here,  the former resembles a relatively small G-JOBASRP instance and the latter a relatively large G-JOBASRP instance.

\begin{figure}[h!]
\centering
\includegraphics[scale=0.75]{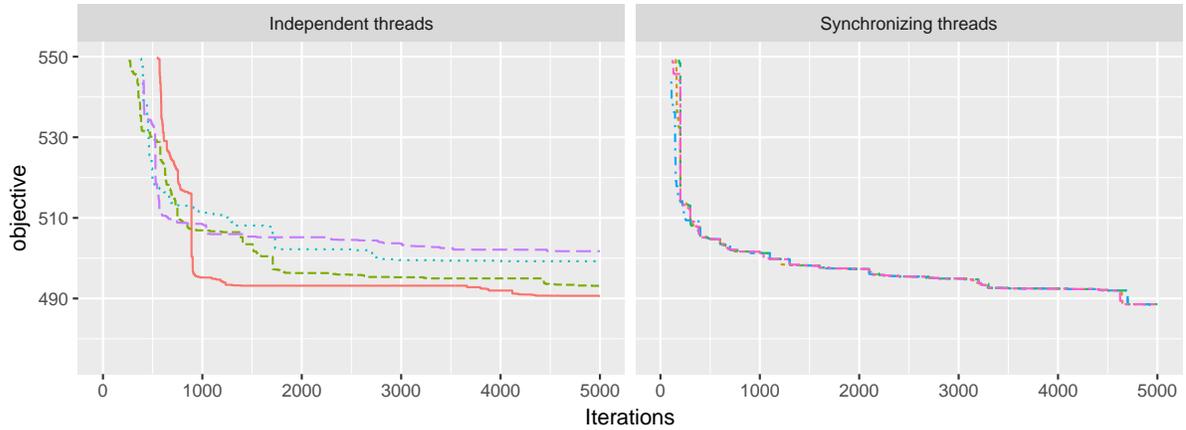}
\caption{Solution trajectory of the ALNS on dataset 6 with 8 order pickers with four threads exchanging information (on the right) and four threads working indepedently (on the left). Each line in each figure corresponds to the current best solution found by a single thread.}
\label{fig:par2}
\end{figure}

In Figures \ref{fig:par1} and \ref{fig:par2}, we depict two example trajectories of the best-known solution found while solving these particular instances. The graphs show that using independent threads result in widely dispersed outcomes. For instance, in Figure 2, the final objectives range between 71 and 79 for the independent threads, while the synchronized threads result in a final objective just above 70. This behavior is the primary benefit of using this simple parallel approach; the variance in the outcomes is reduced sharply, which also results in a faster converging of the best-known solution. 

At the same time, as is advocated by the results in Figure 3 as well, it is observed that the synchronized threads find a better solution than the best solution among the four independent threads. This improvement feels natural, as more computer power will be spent on current solutions with high potential. For instance, the worst-performing thread in Figure 3 still obtained a quarter of the devoted computer power. It is, however, likely that the worst-performing thread is trapped in a local optimum and will not reach a new best-known solution.

Hence, we advocate the use of this parallel technique in any metaheuristic. It does not require more resources than there are readily available at any modern computer, and it is easy to implement. It does, however, reduce the variability and increase the quality of the results.

\subsection{Sensitivity of the order picker's capacity}

\begin{figure}[h!]
\centering
\includegraphics[scale=0.75]{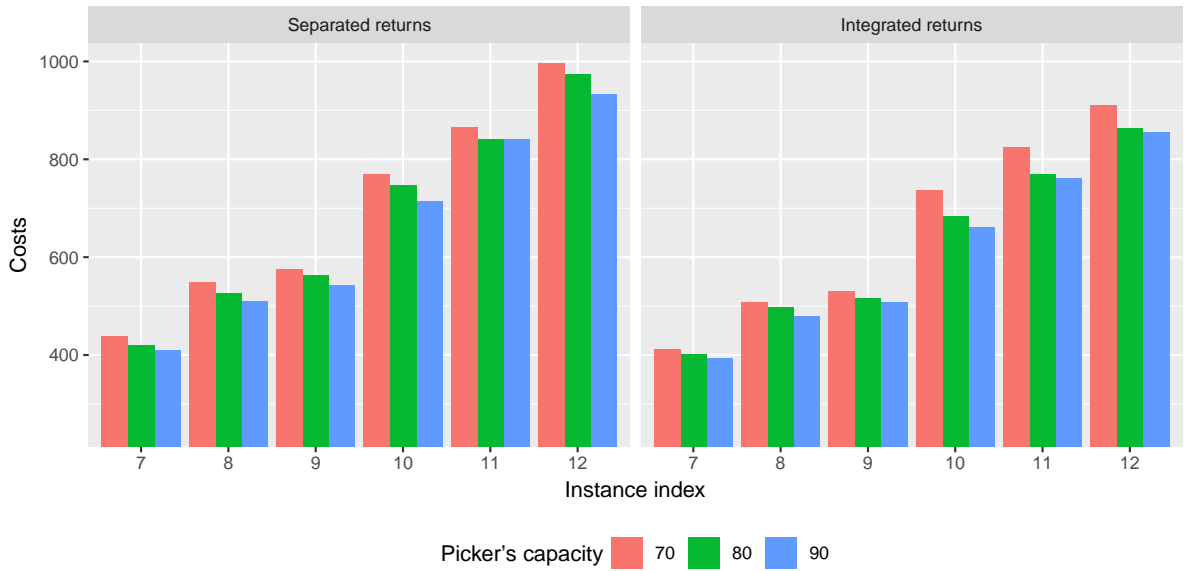}
\caption{Comparison of total costs of the first JOBASPR instances of datasets 7-12 for varying capacity levels. On the left, when product returns are not integrated. On the right, when product returns are integrated.}
\label{fig:bar_cap}
\end{figure}
We furthermore study the effect of reducing and increasing the capacity of the order pickers by 10, and its impact on the integration of product returns in the regular order picking process. In Figure \ref{fig:bar_cap}, we depict for the first G-JOBASPR instances of Datasets 7-12, the effect of increasing picker capacity both when product returns are integrated (on the left) and when products returns are handled separately. 

First, the total costs decrease (slightly) when the picker capacity increases, which is as expected. However, it appears that increasing picker capacity is slightly more useful for the larger instances, as can be observed in Figure \ref{fig:bar_cap}. This effect seems more pressing for handling the returned products in separation. 

Second, the on average cost reduction for integrated processing of product returns equals 6.5\%, 8.3\%, and 7.4\% for an order picker capacity of 70, 80 and 90, respectively. Hence, no conclusion can be drawn on the impact of a smaller or larger picker capacity on the efficiency of handling the product returns integrated with the regular order processing.

\subsection{Impact of the order split-up costs}\label{sec:sens_split}
\begin{figure}
\includegraphics[scale=0.8]{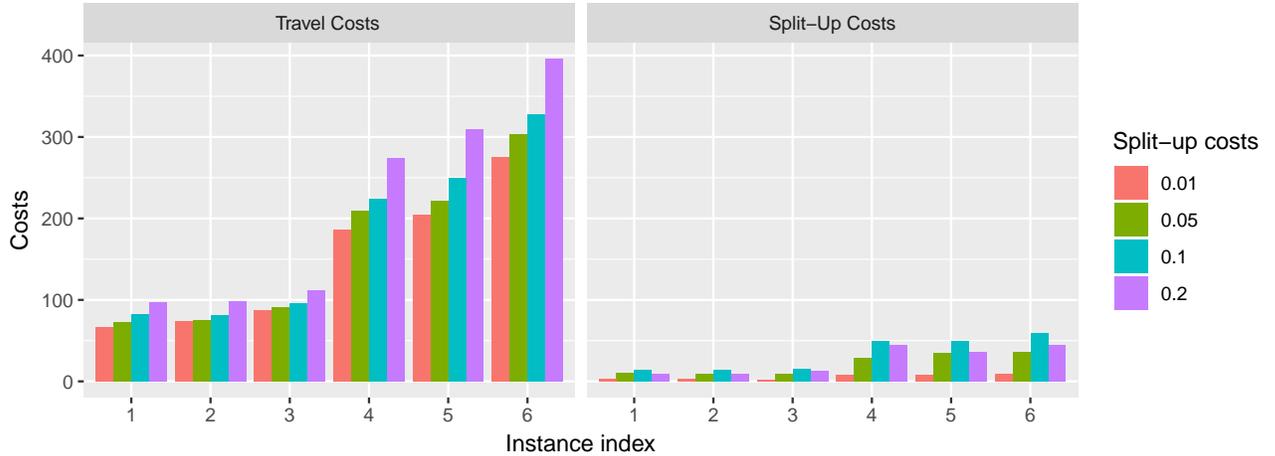}
\caption{The travel and split-up costs for different values of the split-up costs ($\beta$). The instance indices 1-6 refer to the first G-JOBASRP instances of dataset 1-6 respectively}
\label{fig:split-up}
\end{figure}

In order to provide practical insights under which circumstances it is valuable to organize customer order processing in warehouses with order split-ups, we studied the impact of varying the customer order split-up costs $\beta$ on dataset 1-6, again on the first instances of these datasets only. The results are presented in Figure \ref{fig:split-up}.

What stands out is that split-up costs have a large impact on the total costs of the proposed solutions. However, this is not solely caused by the fraction of costs due to splitting up customer orders. Increasing the order split-up costs result in higher travel costs since orders are increasingly assigned to a single batch if split-up costs increase. We observe in  Figure~ \ref{fig:split-up}, that changing the customer order split-up costs from 0.10 to 0.20 leads to a decrease in the total incurred split-up costs. A sharp increase in travel costs accompanies this.   

As a final note, the fraction of split-up costs compared to the travel costs is relatively small. For small split-up costs, the split-up costs are incurred, but they do not form a significant fraction of the total costs. When split-up costs are large, solutions tend to not split-up customer orders, resulting again in a relatively small share of the total costs associated to splitting up customer orders. Concluding, splitting-up customer orders amongst multiple batches have a significant but indirect effect on the total efficiency.

\subsection{Cost partitioning with tight deadlines}

\begin{figure}
\centering
\includegraphics[scale=0.8]{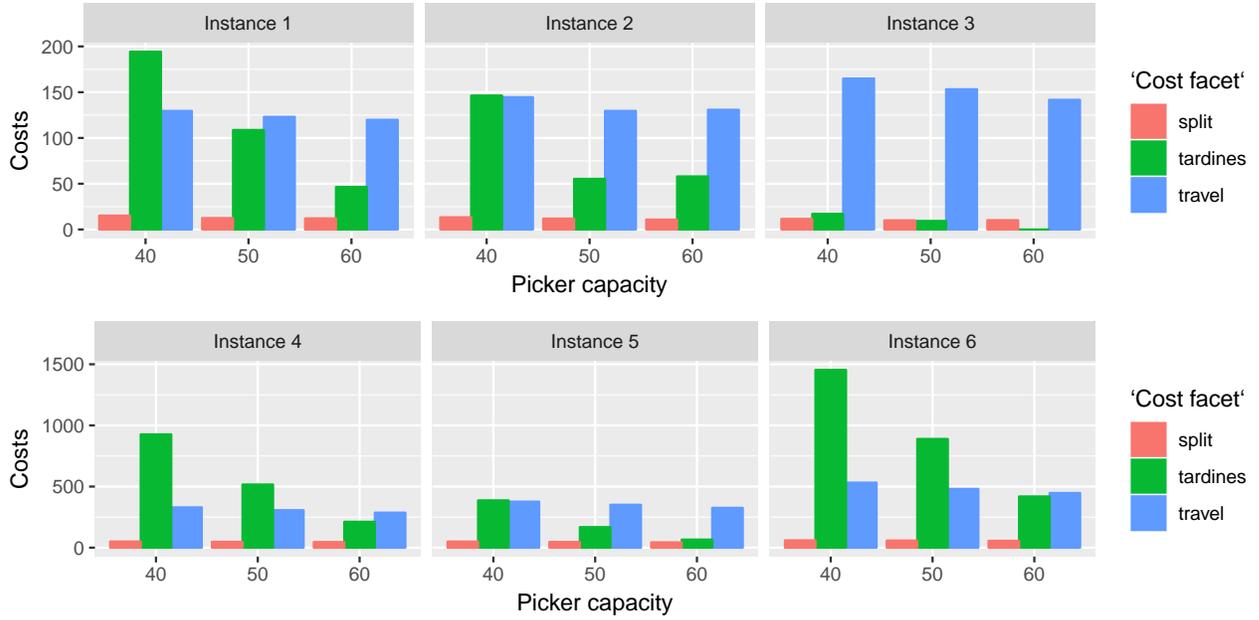}
\caption{The devision of costs caused by travelling, tardines and split-up costs in for the first instances of Datasets 1-6. Customer order deadlines are cut into half and the pickers capacity is varied between 40, 50 and 60.}
\label{fig:bar_deadline}
\end{figure}

The previous analyses are performed in settings where deadline violations did not occur. This is a practically realistic scenario, as violating deadlines might disrupt operations further downstream the supply chain. However, we are interested in the performance of the ALNS and the cost composition of the proposed solutions, in case customer order deadlines are violated. We, therefore, solved the first instances of datasets 1-6 where we cut the deadlines into half and reduced the order picker capacity significantly.

In Figure \ref{fig:bar_deadline}, we present for a varying picker capacity, the division in travel, tardiness, and split-up costs. Several observations are made. First, increasing picker capacity leads to less customer order violations.  Tardiness costs are then reduced significantly while the travel costs and split-up costs remain similar. Second, the split up costs are a fairly constant factor for each instance, and each order picker's capacity. Also, large variations in cost division are observed among the different instances. For example, Instance 3 does almost completely consist of travel costs, whereas instances 1 and 6 are dominated by tardines costs for a capacity of 40. Hence, from these experiments we can conclude that allowing for a sufficient order picker capacity can have a great impact on reducing the tardiness costs. In particular, for practical systems one shouls imply design the warehouse so that order capicty is not a bottleneck of the total system performance.

\subsection{Impact of the MIP operators and results verification} \label{mipop}
We test the impact of using the additional MIP operators as described in Section 4.4. We performed two rounds of additional experiments which we briefly summarize in the following. 

First, it should be noted that the  MIP operators are especially useful to find a good composition of already discovered batches. Hence, we found no additional benefit for using the proposed MIP operators on the datasets and associated G-JOBASRP instances as initially proposed. We, therefore, solved the G-JOBASRP instances associated with datasets 1-6 with a reduced picker capacity to imitate scenarios in which tardiness is observed in the solution. 

We found that, especially in the early stages of the ALNS, a faster converging algorithm is found as the MIP operators found significant improvements by rescheduling the already observed batches. Hence, when less iterations are required, or if any guarentee on the optimality of the proposed schedule is needed (given the batches as discovered by the ALNS), we suggest to include the MIP operators in the ALNS. If, on the other hand, the runtime is not a limiting factor, the solutions proposed by the ALNS with and without MIP operators have a similar performance. 

The second series of experiments are as follow. We solved the associated G-JOBASRP instances of datasets 1-6 with the ALNS without using the MIP operators. While doing so, we store all discovered batches  to suffice as input for the MIP formulations, and accordingly the MIP operators. 

We randomly selected $n=10000$ batches from all the stored batches. We then compared the LP relaxation of the MIP formulations subject to the $10000$ randomly selected batches to the final solution. This is repeated 100 times for all the first instances of Datasets 1-6 with a order picker's capacity of 30 and 40. We did not find any improvement to the final solutions proposed by the ALNS. This shows that the ALNS has no problems scheduling batches correctly. 

Concluding, although the above presented results are no conclusive prove that the ALNS provide near-optimal solutions, it shows that the ALNS provides high-quality solution so that the dynamics of the G-JOBASRP can be analyzed correctly. Lower bounds based on Langrangian relaxation can be constructed to provide an upper bound on the optimality gap, though this is outside the scope of this paper.

\section{Conclusions}
\label{sec:conclusions}
In this paper, we studied the generalized joint order batching, assignment, sequencing and routing problem in picker-to-parts warehouses. The generalizion stems from the inclusion of two characteristics inspired on the recent advances in e-commerce logistics. First, we included the restocking of returned products in the regular customer order processing operations, and second, we investigate the cost-savings potential of picking the distinct order lines that belong to a single customer order amongst multiple order picking batches. 

To formally introduce the problem we studied in this paper, we provided an compact mixed integer programming formulation, as well as two extended formulations. We developed a parallel Adaptive Large Neighborhood Search (ALNS), in which the operators are inspired on classical ALNS moves, on classical iterative local search elements, and searches through neighborhoods defined by the extended MIP formulations. We created a testbed of instances of practical size; the largest solved instances consists of 8038 order lines. 

We compared the performance of the ALNS against two intuitive and practically often observed heuristics and found on-average cost-savings of 17.4\%. Using the ALNS, we show that incorporating product returns in the regular order picking processes can result up to 19\% cost-savings compared to handling the product returns separately. In addition, we tested the maximal cost-savings potential of splitting up customer orders by comparing the situation in which splitting-up is free versus the situation in which splitting-up is not allowed. This is shown to be around 45\%.


In particular, our model and solution algorithm help to organize the order picking process as a whole; to account for the interdependencies between batching and routing, multiple pickers, and deadlines; and to tackle increasing volumes of product returns. We showed that this integrated point of view is especially benificial when warehouses are not working on their full capacity, which is typically observed in practice (except peak-periods around, for instance, Christmas). Hence, research should continue to explore integrated warehouse operation methods. For example, the storage location assignment exerts a substantial impact on order picking performance \citep{theys2010}. Solution procedures with shorter computation times are also desirable as unexpected, short-notice rescheduling might be necessary sometimes. That is, exploiting dynamic strategies might be of interest to pursue. Other optimization models for the order picking problem focussing on determining optimal solutions, or lower bounds to real-life sized instances, are of particular relevance as well. Namely, only if such methods are available it is possible to infer whether more time should be spent on new (meta)heuristic methods. 

\section*{Acknowledgements}
This research is partly funded through grants 438-13-216 (A.H. Schrotenboer) and 439-16-612 (K.J. Roodbergen) of the Netherlands Science Foundation (NWO). 

\section*{References}
\bibliography{batching_routing}
\end{document}